%% file: acl_latex.tex
\pdfoutput=1

\documentclass[11pt]{article}

\usepackage[preprint]{acl}

\usepackage{times}
\usepackage{latexsym}

\usepackage[T1]{fontenc}

\usepackage[utf8]{inputenc}

\usepackage{microtype}

\usepackage{inconsolata}

\usepackage{graphicx}
\usepackage{pifont}
\usepackage{booktabs}
\usepackage{multirow}
\usepackage{CJKutf8}

\usepackage{hyperref}
\usepackage{xcolor}
\usepackage{amsmath}
\usepackage{amssymb}
\usepackage{algorithm}
\usepackage{enumerate}
\usepackage{enumitem}
\usepackage{algpseudocode}
\usepackage{forest}
\usepackage{tikz}
\usepackage{cleveref}
\usetikzlibrary{arrows.meta,shapes,positioning,shadows,trees}
\usepackage{subfigure}
\title{Context-Aware Hierarchical Taxonomy Generation for Scientific Papers via LLM-Guided Multi-Aspect Clustering}

\author{Kun Zhu$^{1}$\thanks{~~Work was done during an internship at SMU.}, Lizi Liao$^{2}$, Yuxuan Gu$^{1}$, Lei Huang$^{1}$, Xiaocheng Feng$^{1,3}$\thanks{~~Corresponding Author}, Bing Qin$^{1,3}$\\
  $^{1}$Harbin Institute of Technology\quad \quad  $^{2}$ Singapore Management University\\
$^{3}$ Peng Cheng Laboratory \\
  \texttt{\{kzhu,yxgu,lhuang,xcfeng,qinb\footnotemark[2]\}@ir.hit.edu.cn} 
  \texttt{ lzliao@smu.edu.sg}}

\begin{document}
\maketitle
\begin{abstract}
The rapid growth of scientific literature demands efficient methods to organize and synthesize research findings. Existing taxonomy construction methods, leveraging unsupervised clustering or direct prompting of large language models (LLMs), often lack coherence and granularity. We propose a novel \textbf{context-aware hierarchical taxonomy generation framework} that integrates LLM-guided multi-aspect encoding with dynamic clustering. 
Our method leverages LLMs to identify key aspects of each paper (\textit{e.g.}, methodology, dataset, evaluation) and generates aspect-specific paper summaries, which are then encoded and clustered along each aspect to form a coherent hierarchy. In addition, we introduce a new benchmark of $156$ expert-crafted taxonomies encompassing $11.6\,\mathrm{k}$ papers, providing the first \textbf{naturally annotated} dataset for this task. Experimental results demonstrate that our method significantly outperforms prior approaches, achieving state-of-the-art performance in taxonomy coherence, granularity, and interpretability.
\footnote{Code and dataset are available in \url{https://github.com/zhukun1020/TaxoBench-CS}.}
\end{abstract}

\input{intro}

\input{method}

\input{experiment}

\input{related}
\section{Conclusion}
In this work, we propose a novel framework for taxonomy generation that leverages multi-dimensional representations and dynamic clustering. By dynamically generating semantic aspects tailored to each document set and searching for optimal clustering configurations via dynamic search, our method constructs taxonomies that are both semantically coherent and structurally faithful. We further introduce a high-quality benchmark of 156 annotated taxonomies derived from CS survey papers to facilitate reliable evaluation.
Extensive experiments demonstrate that our approach outperforms existing pure LLM-based and clustering-incorporated methods in both automatic and human evaluations. Ablation studies confirm the effectiveness of dynamic aspect modeling and adaptive clustering strategies.

\input{limitation}

\bibliography{custom}

\clearpage
\newpage
\appendix
\input{appendix}

\end{document}

%% file: intro.tex
\section{Introduction}

The rapid expansion of academic publications has created an overwhelming amount of information, making it increasingly challenging for researchers to stay up-to-date and systematically organize domain knowledge \cite{Reisz_2022,10.1162/qss_a_00327,vineis2024scientific}. 
As a result, there is a growing demand for structured and concise taxonomies that can support the exploration and synthesis of more efficient literature \cite{shen_HiExpanTaskGuidedTaxonomy_2018,zhu-etal-2023-hierarchical}.
Traditional approaches to building taxonomies of scientific papers typically rely on manual or narrowly defined schemes. Common solutions include supervised classification into a predefined hierarchy (\textit{e.g.}, ACM CCS) \cite{zhang2021hierarchical, sadat-caragea-2022-hierarchical, rao2023hierarchical} and unsupervised clustering of papers followed by post-hoc keyword-based label extraction \cite{zhang_TaxoGenUnsupervisedTopic_2018a, shang_NetTaxoAutomatedTopic_2020}. These methods often require substantial human curation or yield coarse topic structures, limiting their usefulness for in-depth literature understanding.

\begin{figure}[t]
    \centering
    \includegraphics[width=1\linewidth]{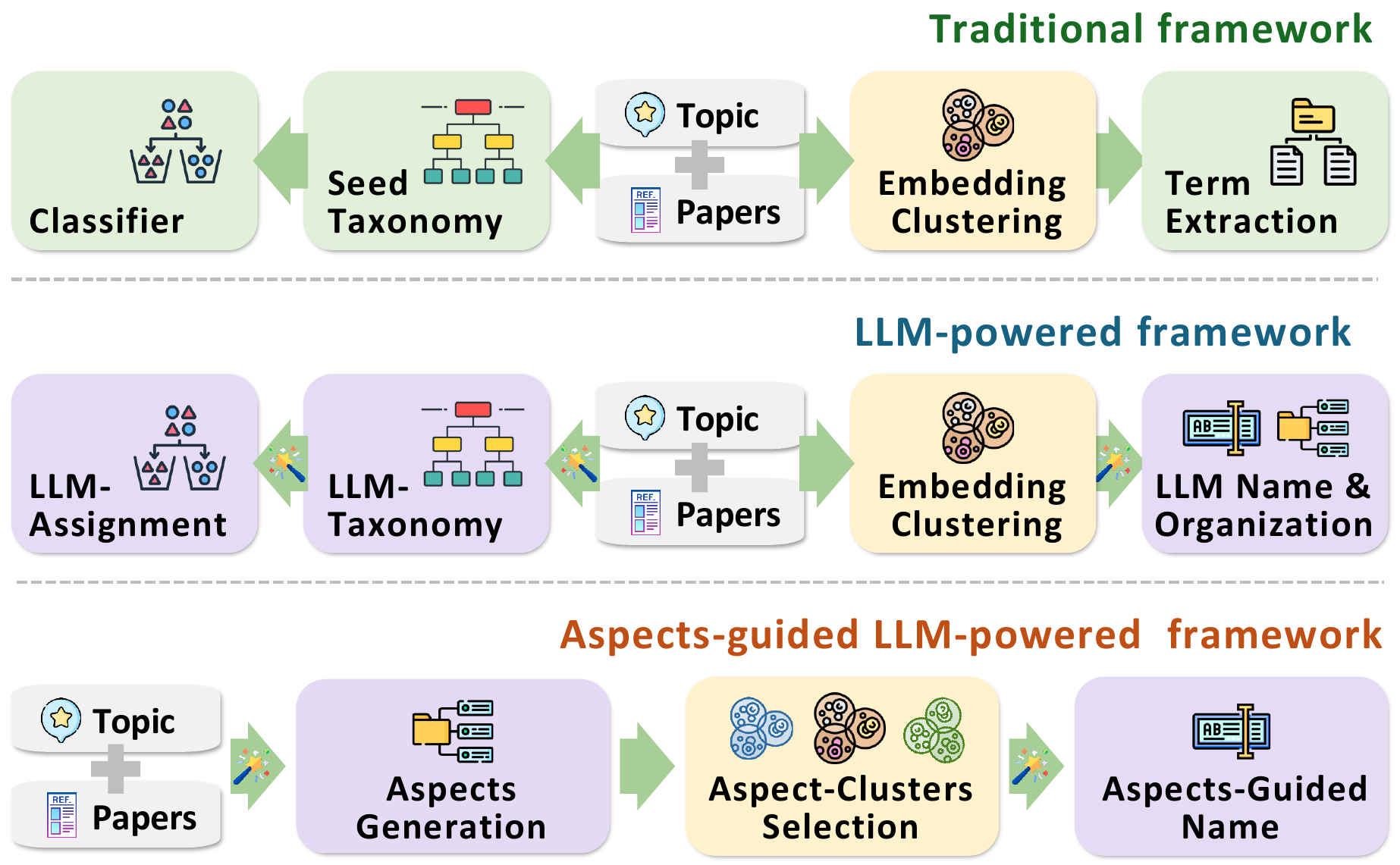}
    \caption{Comparison of taxonomy construction paradigms. Traditional methods typically use supervised classification or clustering with term extraction. Recent approaches incorporate LLMs to replace or enhance key components within these pipelines (purple). Our approach uniquely integrates LLMs with clustering in a context-aware multi-aspect framework, resulting in coherent and precise hierarchical taxonomies.}
    \vspace{-0.2cm}
    \label{fig:intro}
\end{figure}

Recent advances utilize LLMs to automate the taxonomy construction. LLMs demonstrate strong capabilities in long-text understanding and abstraction \cite{achiam2023gpt,grattafiori2024llama}, leading to approaches that generate taxonomy trees or assign papers to categories in an end-to-end fashion \cite{hsu2024chime,wan_TnTLLMTextMining_2024}. 
Hybrid strategies first cluster papers and then prompt LLMs to produce summaries or category labels for each cluster \cite{katz-etal-2024-knowledge, hu_TaxonomyTreeGeneration_2025}.

While these LLM-based methods have shown promise, studies have found that they struggle to capture highly specialized knowledge and fine-grained concepts specific to scientific domains. Moreover, taxonomies produced solely by LLMs are not guaranteed to align with the content of a given corpus, often resulting in missing or hallucinated categories.
Effective taxonomy construction inherently demands context-aware representations, wherein the characterization of each paper dynamically adapts based on its relationships and similarities to surrounding papers. Without this context awareness, papers focusing on distinct aspects (\textit{e.g.}, methodologies \textit{v.s.} datasets) might be incorrectly categorized, leading to incoherent taxonomy structures. This gap calls for new techniques that consider multiple content dimensions and their corpus-level context during taxonomy generation.

In this paper, we propose a novel framework for paper taxonomy generation that leverages LLM-guided, multi-aspect representations in conjunction with adaptive clustering. Specifically, our approach uses a dynamic aspect generator to automatically determine salient semantic aspects (such as research objective, methodology, or data source) for a given collection of papers. Guided by these, the LLM produces aspect-specific summaries for each paper, ensuring that each document is represented in a manner that is both facet-specific and context-aware. We then employ a dynamic clustering algorithm to search for an optimal grouping of papers for each aspect dimension. By iteratively applying multi-aspect encoding and clustering in a top-down fashion, our framework constructs a hierarchical taxonomy tree that is tailored to the corpus at each level. This design allows the taxonomy to capture different facets of the literature at different branches, yielding more coherent and interpretable category structures.

In addition to methodological innovations, a significant bottleneck in this area has been the lack of high-quality, naturally annotated datasets for evaluating taxonomy construction. Most existing benchmarks are synthetic \cite{hsu2024chime} or rely on coarse \cite{katz-etal-2024-knowledge}, predefined categories that fail to reflect the nuanced hierarchies. To bridge this gap, we construct a new dataset of academic taxonomies \textbf{TaxoBench-CS}, by collecting $156$ human-authored taxonomy trees (covering $11.6\,\mathrm{k}$ research papers) from survey and review articles on \texttt{arXiv}. These taxonomies, created by domain experts, provide realistic hierarchical structures that mirror a deep understanding of topic decomposition. This dataset offers a valuable resource for training and evaluating taxonomy generation methods under more natural conditions, and we will release it to foster further research.

In summary, our contributions are threefold:
\begin{itemize}[topsep=-1pt]
\setlength{\parskip}{1pt}
\setlength{\itemsep}{0pt plus 1pt}
    \item We curate a high-quality benchmark consisting of $156$ expert-annotated taxonomies of $11.6\,\mathrm{k}$ papers, facilitating future research. 
    \item We propose to combine multi-aspect paper encoding with a dynamic clustering algorithm, enabling context-aware, hierarchical organization of research papers. 
    \item Our approach outperforms existing state-of-the-art methods, yielding interpretable and human-readable taxonomy trees with significantly improved coherence and granularity.
\end{itemize}



%% file: method.tex
\begin{table*}[t]
  \centering
  \small
    \begin{tabular}{ccccc}
    \toprule
    Datasets & Clustering & Hierachy & Ground Truth & Source\\
    \midrule
    CLUSTREC-COVID \cite{katz-etal-2024-knowledge} & \textcolor{green!80!black}{\ding{51}} & \textcolor{red!90!black}{\ding{55}}  & \textcolor{green!80!black}{\ding{51}} & \textcolor{red!90!black}{\texttt{synthetic}} \\
    SCITOC \cite{katz-etal-2024-knowledge} & \textcolor{red!90!black}{\ding{55}}  & \textcolor{green!80!black}{\ding{51}} & \textcolor{green!80!black}{\ding{51}} & \textcolor{green!80!black}{\texttt{natural}} \\
    SciPile \cite{gao_ScienceHierarchographyHierarchical_2025} & \textcolor{green!80!black}{\ding{51}} & \textcolor{green!80!black}{\ding{51}} & \textcolor{red!90!black}{\ding{55}}  & \textcolor{red!90!black}{\texttt{synthetic}} \\
    CHIME \cite{hsu2024chime} & \textcolor{green!80!black}{\ding{51}} & \textcolor{green!80!black}{\ding{51}} & \textcolor{red!90!black}{\ding{55}}  & \textcolor{red!90!black}{\texttt{synthetic}} \\
    TaxoBench-CS (Ours) & \textcolor{green!80!black}{\ding{51}} & \textcolor{green!80!black}{\ding{51}} & \textcolor{green!80!black}{\ding{51}} & \textcolor{green!80!black}{\texttt{natural}} \\
    \bottomrule
    \end{tabular}%
  \caption{Comparison of existing taxonomy datasets: Datasets are evaluated based on three key criteria: clustering annotations, hierarchical structures, and ground-truth labels. We also distinguish whether datasets are synthetic or naturally derived. Our dataset uniquely meets all three criteria while being naturally sourced. }
  \vspace{-0.2cm}
  \label{tab:dataset}%
\end{table*}%

\section{Preliminary}
Here, we first formalize the task of taxonomy construction for scientific literature. We then describe the creation of a new benchmark dataset derived from human-authored taxonomies in survey papers.

\subsection{Task Definition}

Given a specific topic $x$ and a collection of corresponding scientific papers $\mathcal{D} = \{d_1, d_2, \dots, d_N\}$, the objective is to generate a hierarchical taxonomy $\mathcal{T}(V,E)$ that organizes these papers into a tree structure of semantically coherent categories. In detail, the taxonomy of depth $L$ starts from a root node $r\in V^{(0)}$ and each node $v\in V^{(l)}$ corresponds to a depth $l$, where $V=\bigcup\nolimits_{l=0}^L V^{(l)}$. In addition, each node $v$ is associated with a subset of papers $D_v\subseteq \mathcal{D}$ and a topic facet $x_v$ (\textit{e.g.}, high-level methodological approaches, underlying mechanisms or learning paradigms, or specific research tasks and evaluation scenarios).
The root node $r$ represents the overarching topic $x$ and encompasses all papers $D_r = \mathcal{D}$.
For every non-leaf node $v \in V^{(l<L)}$, its $k_v$ child nodes $\text{Child}(v)$ form a complete, non-overlapping partition of the papers subset $D_v$, satisfying the constraints:
\begin{equation}
    \begin{aligned}
    &\text{Child}(v)=\big\{v_1, v_2,\dots,v_{k_v}\big\}\subseteq V^{(l+1)},\\ &\quad\text{with}\left\{\begin{aligned}&\qquad\bigcup\nolimits_{t=1}^{k_v} D_{v_t} = D_v\\&D_{v_t}\bigcap D_{v_{t'}}=\emptyset,\; \forall t \neq t'
    \end{aligned}\right..
    \end{aligned}
\end{equation}
Edges typically represent hierarchical semantic relations (\textit{e.g.}, \textit{isA}, \textit{instanceOf}) and are restricted to link nodes across adjacent layers, where
\begin{equation}
    E = \bigcup\nolimits_{l=0}^{L-1} E^{(l)},\ E^{(l)}\subseteq V^{(l)}\times V^{(l+1)}.
\end{equation}
In our framework, the taxonomy is built iteratively by partitioning each subset $D_v$ from the depth $l$ into disjoint subsets assigned to its children.

\subsection{Dataset Construction}

Existing datasets for evaluating taxonomy generation methods generally rely on either topic-based retrieval followed by manual annotation \cite{katz-etal-2024-knowledge} or LLM-assisted taxonomy creation and filtering \cite{hsu2024chime,gao_ScienceHierarchographyHierarchical_2025}. However, these approaches often introduce noise into the structure and lack high-quality, reliably annotated ground-truth hierarchies.

To address these limitations, we introduce a new benchmark dataset, \textbf{TaxoBench-CS}, constructed from naturally annotated taxonomy trees found in computer science review papers on \texttt{arXiv}\footnote{\href{https://arxiv.org/}{https://arxiv.org/}}. We start by systematically selecting survey papers that contain explicit hierarchical taxonomy diagrams. By parsing the corresponding \LaTeX\ source files, we extract citation identifiers directly linked to taxonomy structures, which are then mapped to their full titles using the citation metadata provided in each paper’s associated \texttt{.bib} or \texttt{.bbl} files. Next, we retrieve detailed paper metadata from Semantic Scholar\footnote{\href{https://www.semanticscholar.org/me/research}{https://www.semanticscholar.org/me/research}}. To ensure the dataset’s accuracy and reliability, we manually verify all citation mappings, eliminating any incorrect or ambiguous entries.


The final TaxoBench-CS dataset consists of $156$ author-curated taxonomy trees, serving as robust hierarchical annotations. Each taxonomy contains, on average, $74.4$ referenced papers and spans $3.1$ levels in depth. Excluding the paper citation indicators connected to the leaf-level nodes, each tree includes around $24.8$ nodes that represent structured semantic categories, providing a rich and structurally sound resource. 
As shown in Table \ref{tab:dataset}, our proposed TaxoBench-CS uniquely combines explicit clustering structures, hierarchical organization, and authoritative annotations derived directly from naturally occurring expert-curated taxonomies. This combination makes it an ideal benchmark for evaluating and developing taxonomy generation methods under realistic conditions. 

\begin{figure*}[t]
    \centering
    \includegraphics[width=1\linewidth]{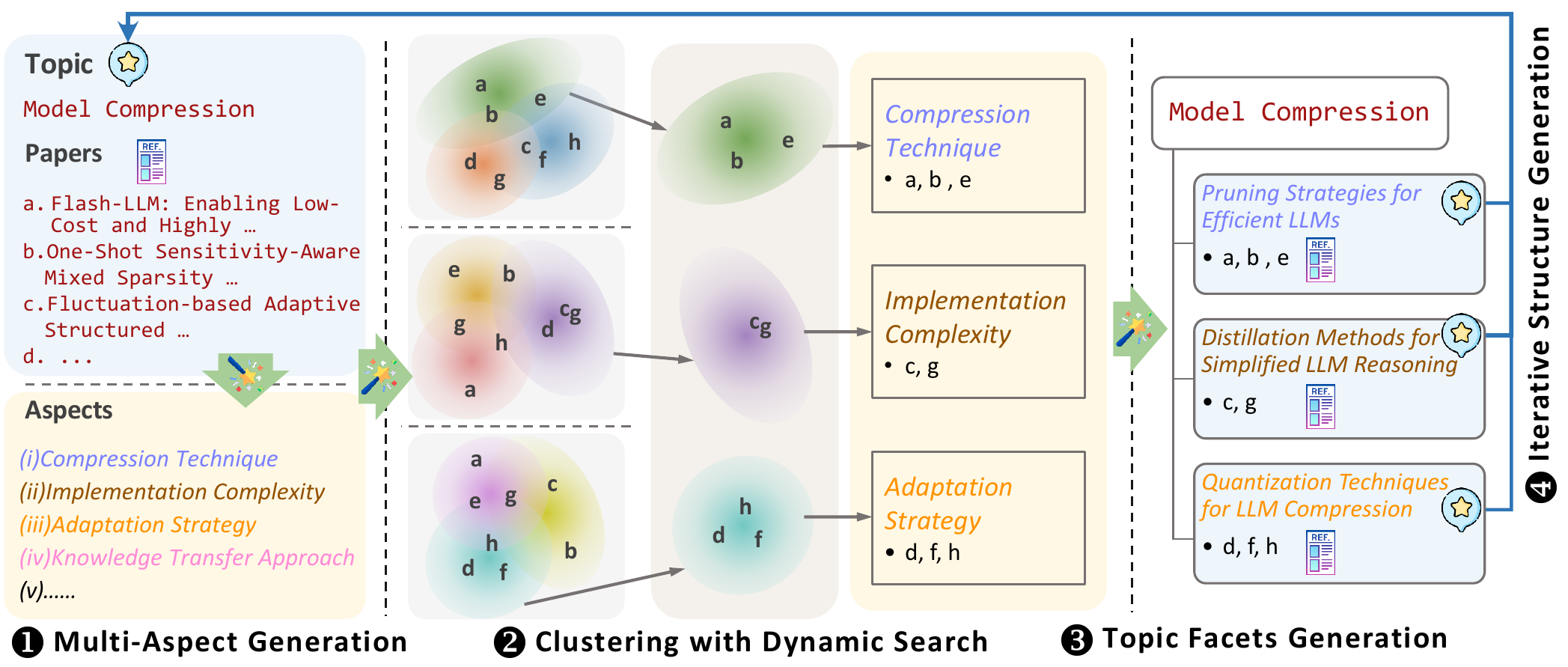}
    \caption{
    Our proposed Aspects-guided LLM-based Top-Down Clustering framework. 
    Specifically, we dynamically generate multiple semantic aspects to represent each paper, and perform aspect-specific clustering via dynamic search. The abstract aspects are instantiated into concrete topic facets, which serves as the heading of nodes. This process is iteratively applied to construct a coherent and semantically meaningful taxonomy.}
    \label{fig:method}
\end{figure*}

\section{Method}

The core of our method lies in appropriately decomposing the given node $v$ of depth $l$ according to the structure and semantics of its associated paper set $D_v$. We first represent papers in the associated paper set $d_i\in D_v$ using multi-aspect encoding (\S \ref{sec:encoding}). Given the clustering results over the multi-aspect vectors of $D_v$, we apply a dynamic search algorithm to determine the most appropriate partitioning strategy (\S \ref{sec:clustering}). 
Therefore, we can iteratively partition the paper set $D_v$ and get the child nodes $\text{Child}(v)$ of node $v$ from a top-down manner to construct the taxonomy tree (\S \ref{sec:iterative}).

\subsection{Multi-Aspect Paper Encoding}
\label{sec:encoding}

In this part, our goal is to obtain a global representation of the paper set $D_v$ that captures its overall semantic structure. To this end, we propose to automatically generate a set of candidate aspects $\mathcal{A}_v$ using an LLM based on all papers in $D_v$. These aspects are then used in a parallel manner to guide the encoding of individual papers. The aspect generator is defined as follows:
\begin{equation}
    \label{eq:aspect}
    \mathcal{A}_v \sim p_\text{LLM}(\mathcal{A}|v,D_v),
\end{equation}
where we prompt the LLM such as GPT-4o to analyze the paper distribution in $D_v$ according to the global trace of current node $v$ (topic facets of $v$ and all its ancestor nodes) before generating the detailed content of aspects $\mathcal{A}_v$. In addition, the LLM is required to infer the number of aspects $\vert\mathcal{A}_v\vert$ automatically.
We demand the LLM to identify a set of salient semantic dimensions that can effectively characterize and classify the papers, such as research problem and application domain.

Given the discovered aspects $a\in\mathcal{A}_v$, we parallelly generate aspect-guided summaries $s^d_a$ for each paper $d\in D_v$ by prompting the LLM. Each summary is then encoded into a $n$-dimensional vector $e^d_a\in\mathbb{R}^n$, where we have:
\begin{equation}
    \label{eq:summary}
    \begin{aligned}
        &\textit{For all} \ \ a, d\  \in \mathcal{A}_v \times D_v\ \textit{in parallel}:\\
        &\quad\quad e^d_a = \text{Enc}(s^d_a),\quad s^d_a \sim p_\text{LLM}(s|a,d)\ .
    \end{aligned}
\end{equation}
We collect the encoding of paper set $D_v$ for each aspect and obtain $\mathbf{e}_a=\{e^d_a \  \vert \ \forall d \in \mathcal{D}_v\}$, which can also be regarded as a matrix $\mathbf{e}_a\in\mathbb{R}^{\vert D_v\vert \times n}$.



\subsection{Clustering with Dynamic Search}
\label{sec:clustering}
Given that encoding across different aspects may reside in heterogeneous semantic spaces with varying structures and scales, directly aggregating all representation vectors $\mathbf{e}=\{e_a^d\ \vert\ \forall d\in D_v,\forall a\in\mathcal{A}_v\}$ into a unified space for clustering would be inappropriate. Therefore, we perform clustering independently within each aspect space $\mathbf{e}_a$:
\begin{equation}
    \label{eq:cluster}
    \notag
    \begin{aligned}
        &\textit{For all} \ \ a\  \in \mathcal{A}_v\ \textit{in parallel}:\\
        &\quad f_a:\mathbf{e}_a\times\{1,2,\dots,k\}\rightarrow [0,1]\\
        &\quad\text{Expectation}:\; \forall i \in \{1,2,\dots,k\},\\
        &\qquad \mathcal{C}^i_a\!=\!\big\{e_a^d \ \big\vert\  \arg \max_j f_a(e_a^d, j) = i,\forall d\in D_v\big\}\\
        &\quad \text{Maximization}:\\
        &\qquad \mathcal{L}^\text{cluster}_a = -\sum\nolimits_{i=1}^k\sum\nolimits_{e\in \mathcal{C}^i_a} f_a(e,i),
    \end{aligned}
\end{equation}
\noindent\hfill\smash{\raisebox{20pt}{(\refstepcounter{equation}\theequation)}}\par
\noindent where $\mathcal{C}_a^i$ is the temporary allocation of the cluster index $i$ and $f_a$ is the clustering model that maps the encoding vector $e$ to the cluster $i$ with a probability of $f_a(e,i)$, $\sum_{i=1}^{k}f_a(e,i)=1$. In addition, $k$ is a hyperparameter that determines the number of clusters, where $k_v \leq |\mathcal{A}_v|\times k$.

Given the cluster assignment probabilities for each aspect, we need to select for each paper $d\in D_v$ a unique pair $(a,i)$, where $a$ is an aspect and $i$ is a cluster index within that aspect, such that: (1) Each paper $d$ will be assigned to only one cluster $i$. (2) The total number of unique pairs $(a,i)$ used in the paper set $D_v$ is $k_v$. (3) The total assignment probability is maximized. Therefore, we define a binary indicator $\delta^{d}_{a,i} \in \{0, 1\}$ and the objective:
\begin{equation}
    \max_{\delta} \sum_{d \in D_v} \sum_{a \in \mathcal{A}_v} \sum_{i=1}^{k} \delta^{d}_{a,i} \cdot f_a(e_a^d, i),
\end{equation}
which is subject to:
\begin{equation}
    \begin{aligned}
    &\qquad\sum_{a \in \mathcal{A}_v} \sum_{i=1}^{k} \delta^{d}_{a,i} = 1, \forall d \in D_v\\
    &\left| \left\{ (a, i)\ \big\vert\ \exists d \in D_v \text{ s.t. } \delta^{d}_{a,i} = 1 \right\} \right| = k_v.
    \end{aligned}
\end{equation}
\begin{algorithm}[t]
\small
\caption{Search \textcolor{gray!90!white}{with Pruning}}
\label{alg:search}
\begin{algorithmic}[1]
\State \textbf{Init. } $\mathbb{S}\!\gets\!\left\{S\!\subseteq\!\mathcal{A}_v\!\times\!\{1,\dots,k\}\,\vert\,  \vert S\vert\!=\!k_v\right\}$
\State \textbf{Init. } $\texttt{score}[S] \gets 0, \; \forall S \in \mathbb{S}$
\State \textbf{Init. } $\texttt{state}[S][(a,i)]\gets \{\}, \; \forall S \in \mathbb{S},\ (a,i)\in S$
\ForAll{$d\in D_v$ \textcolor{gray!90!white}{in random order}}
    \ForAll{$S \in \mathbb{S}$}
                \State $\texttt{score}[S]\!\gets\! \texttt{score}[S]\!+\!\max\limits_{(a,i)\in S} f_a(e^d_a,i)$
                \State $\texttt{state}[S][\arg\max\limits_{(a,i)\in S}f_a(e^d_a,i)].add(d)$
    \EndFor
    \textcolor{gray!90!white}{\If{$\text{score}[S]\ll \text{avg}\ \texttt{score},\ \exists S\in\mathbb{S}$}
    \State $\mathbb{S} \gets \mathbb{S} \setminus S$
    \EndIf}
\EndFor
\State $\texttt{max\_score}\gets\max\limits_{S\in \mathbb{S}}\,\texttt{score}[S]$
\State $S^*\gets \arg\max\limits_{S\in\mathbb{S}}\,\texttt{score}[S]$
\State $\texttt{optimal\_state}\gets\texttt{state}[S^*]$
\State \Return $S^*,\ \texttt{max\_score},\  \texttt{optimal\_state}$
\end{algorithmic}
\end{algorithm}
As a result, we have the search process as illustrated in the algorithm \ref{alg:search}, where we directly define a search space $\mathbb{S}$ containing all possible combinations $S \subseteq \mathcal{A}_v \times \{1,2, \dots, k\}$ that satisfy $|S| = k_v$.
Each $S$ encodes a specific clustering scheme with $k_v$ unique aspect-cluster assignments $(a,i)$. We adopt a real-time strategy that the \texttt{score} of every combination $S$ is updated as each paper $d \in D_v$ arrives, where we trace the optimal assignment trajectory via the \texttt{state} variable.
Optionally, we can randomize the iterative order of the papers and prune the low-scoring combinations during the process to reduce search space and improve efficiency. After processing all documents, the algorithm returns the highest score combination $S^*$ along with its trajectory \texttt{optimal\_state}.

We can extract the partitioned paper sets $D_{v_t}$ from the trajectory \texttt{optimal\_state} and generate the topic facet $x_{v_t}$ with LLM as follows:
\begin{equation}
    \label{eq:facet}
    \notag
    \begin{aligned}
        &\textit{For all} \ \ (a,i) \in S^*,\ t\in \{1,\dots,k_v\}\ \ \textit{in parallel}:\\
        &\quad D_{v_t} = \{d\mid \forall d \in\texttt{optimal\_state}[(a,i)]\}\\
        &\quad  x_{v_t} \sim p_\text{LLM}(x|v,D_{v_t},S^*)\\
        &\quad v_t \triangleq \langle x_{v_t}, D_{v_t} \rangle,\ E^{(l)} \gets E^{(l)} \cup \{(v, v_t)\},
    \end{aligned}
\end{equation}
\noindent\hfill\smash{\raisebox{30pt}{(\refstepcounter{equation}\theequation)}}\par
\vspace{-0.3cm}
\noindent where the node $v_t$ is connected to its parent $v$.



\subsection{Iterative Structure Generation}
\label{sec:iterative}

As illustrated in Figure \ref{fig:method}, our method constructs the taxonomy in a top-down manner, starting from the root node $r$ and iteratively expanding the child nodes $\text{Child}(v)$ for node $v$ from each depth $l$, this is decomposing the associated paper set $D_v$ and generating a corresponding topic facet $x_v$ that characterizes the semantic focus of its substructure.

During each expansion step, we dynamically generate new aspects based on the current distribution of the papers in $D_v$. 
This process is tailored to capture the updated salient semantic dimensions and key distinctions among papers within the new partitioned subset. 
It is worth noting that we incorporate the topic facets of all ancestor nodes into the prompt context. 
This ensures that the newly generated aspects reflect not only local document features, but also the global structural direction of the taxonomy, thereby better understanding the direction in which the current node needs to be expanded.
The expansion process continues until a stopping condition is met, such as reaching a maximum depth $L$ or encountering the number of papers in the node below a predefined threshold. 
Once the expansion is complete, the resulting tree constitutes the taxonomy of given topic and papers.



%% file: experiment.tex
\section{Experiments}

\subsection{Baselines}

We compare our approach with two categories of methods: pure LLM-based and clustering-incorporated taxonomy generation.

\subsubsection{Pure LLM-based Methods}

\textbf{CHIME} \cite{hsu2024chime} extracts claims and frequent entities from related papers, then prompts an LLM to generate root categories and assign claims into a hierarchical structure. 

\noindent\textbf{TNT-LLM} \cite{wan_TnTLLMTextMining_2024} first prompts an LLM to summarize each input, then iteratively constructs and refines a taxonomy from the summaries. 

\noindent\textbf{GoalEx} \cite{wang2023goal} generates explanation-based candidate clusters given a goal, and assigns each document via entailment prompting. A integer linear programming step selects clusters that best cover the dataset with minimal redundancy.

\subsubsection{Clustering-incorporated Methods}

\textbf{Knowledge Navigator} \cite{katz-etal-2024-knowledge} encodes paper abstracts into dense embeddings and applies traditional clustering algorithms to group them. The resulting clusters are named and organized into a hierarchical structure by LLM.

\noindent\textbf{SCYCHIC} \cite{gao_ScienceHierarchographyHierarchical_2025} uses an LLM to extract structured contributions from each paper, which are then embedded and clustered hierarchically. A bidirectional clustering algorithm specifies the number of levels and clusters per level.

\begin{table*}[t]
  \centering
  \small
    \begin{tabular}{lcccccc|ccccc}
    \toprule
      & \multicolumn{3}{c}{\textbf{Categorization}} & \multicolumn{2}{c}{\textbf{Structure}} & \multicolumn{1}{c|}{\multirow{2}{*}{\textbf{Nodes}}} & \multicolumn{5}{c}{\textbf{Human Assessment}} \\
      & \textbf{NMI} & \textbf{ARI} & \textbf{Purity} & \textbf{CEDS} & \textbf{HSR} &   & \textbf{Cov.} & \textbf{Rel.} & \textbf{Str.} & \textbf{Val.} & \textbf{Ade.}\\
    \midrule
    \multicolumn{12}{l}{\textbf{Pure LLM-based}}  \\
    \midrule
    CHIME & 35.4  & \phantom{0}0.9  & 41.8  & \underline{23.3}  & \textbf{74.7}  & 1.1  & 43.2  & 50.3  & 54.5  & 47.6  & \underline{47.6}  \\
    TnT-LLM & \underline{51.6}  & \phantom{0}2.3  & \underline{57.6}  & 19.1  & 69.9  & 1.5  & 41.1  & 47.3  & 48.1  & 46.0  & 46.6  \\
    GoalEx & 46.7  & \phantom{0}8.8  & 47.6  & 23.2  & 70.5  & 1.0  & 45.9  & 53.3  & \underline{57.0}  & 48.6  & 46.8  \\
    \midrule
    \multicolumn{12}{l}{\textbf{Clustering-incorporated}}\\
    \midrule
    KN & 44.7  & \underline{16.2}  & 42.4  & 18.8  & 49.5  & 0.5  & \underline{47.5}  & \underline{57.0}  & 55.0  & \underline{52.0}  & 47.0  \\
    SCYCHIC & 49.8  &  \phantom{0}9.0 & 50.6  & 23.0  &  66.4 & 1.5  & 47.3  & 50.7  & 55.2  & 48.4  & 46.8 \\
    \midrule
    Ours & \textbf{60.1} & \textbf{19.1} & \textbf{62.2} & \textbf{23.8} & \underline{74.5}  & 1.2  & \textbf{50.6} & \textbf{57.1} & \textbf{59.6} & \textbf{52.9} & \textbf{54.4} \\
    \bottomrule
    \end{tabular}%

  \caption{Automatic and human evaluation results on taxonomy generation. We report categorization quality (\textbf{NMI}, \textbf{ARI}, \textbf{Purity}), structural consistency (\textbf{CEDS}, \textbf{HSR}), and normalized node count (\textbf{Nodes}), where $1.0$ of \textbf{Nodes} indicates an exact match with the gold taxonomy in terms of node count. Human evaluation is conducted on five dimensions, \textbf{Cov}erage, \textbf{Rel}evance, \textbf{Str}ucture, \textbf{Val}idity, and \textbf{Ade}quacy, each rated on a scale of $1$ to $100$.}
   \vspace{-0.2cm}
    \label{tab:main-results}%
\end{table*}%

\subsection{Experimental Settings}

We employ GPT-4o (2024-08-06) for aspect generation (eq.\ \ref{eq:aspect}) and topic facet generation (eq.\ \ref{eq:facet}), due to its superior reasoning and abstraction capabilities.
Besides, we use LLaMA-3.1-8B to generate aspect-guided summaries (eq.\ \ref{eq:summary}), as it requires less complex reasoning to locate and extract relevant information from the paper. 
This division enables a balance between generation quality and computational cost across the pipeline.

Following \citet{katz-etal-2024-knowledge}, we adopt text-embedding-3-large for paper encoding (eq.\ \ref{eq:summary}) and use Gaussian Mixture Models (GMMs) as the aspect-specific clustering model $f_a(e,i)$ (eq.\ \ref{eq:cluster}).
In the main experiments, the number of clusters per aspect $k$ and the number of child nodes per parent node $k_v$ are both empirically set as 4. 
The maximum taxonomy depth is limited to $L=3$. See the prompts that we use in the Appendix~\ref{sec:prompts}.
Due to computational and manual costs, we randomly sample $25$ of the $156$ taxonomy instances for human evaluation and ablation studies. Each configuration was executed once with a fixed random seed, and results are averaged over the sampled instances.

\subsection{Evaluation Metrics}


We evaluate taxonomy generation from two complementary perspectives: \textit{papers categorization} and \textit{topic structure}, using both automatic and human evaluation. Full metric definitions and annotation guidelines are provided in Appendix~\ref{sec:appendix-metrics}.

\noindent \textbf{Automatic Evaluation.}
To assess papers categorization, we report three widely used clustering metrics: Normalized Mutual Information (\textbf{NMI}), Adjusted Rand Index (\textbf{ARI}), and \textbf{Purity}.
For topic quality and structural alignment, we adopt Heading Soft Recall (\textbf{HSR}) \cite{franti2023soft} and Catalogue Edit Distance Similarity (\textbf{CEDS}) \cite{zhu-etal-2023-hierarchical}. 
In addition, we use a normalized \textbf{Nodes Ratio}, defined as the number of generated nodes divided by the number of nodes in the oracle taxonomy, as an auxiliary metric to monitor coarse-grained structural discrepancies.

\noindent \textbf{Human Evaluation.}
Following \citet{hu2024taxonomy}, we conduct human evaluation on five dimensions: \textbf{Coverage}, \textbf{Relevance}, \textbf{Structure}, \textbf{Validity}, and \textbf{Adequacy}. Each dimension is rated on a scale of 1 to 100 to allow fine-grained comparisons. The evaluation is performed by six reviewers: three PhD students in computer science and three advanced LLMs: GPT-4o (2024-11-20), Claude 3.7 Sonnet (2025-02-19), and LLaMA-3.3-70B Instruct. 

\subsection{Main Results}

\textbf{Best categorization performance.}
We obtain the best categorization performance, with NMI ($60.1$), ARI ($19.1$), and Purity ($62.2$), surpassing both pure LLM-based baselines (\textit{e.g.}, TnT-LLM with NMI of $51.6$) and clustering-incorporated baselines (\textit{e.g.}, KN with ARI of $16.2$). 
This proves the superiority of our multi-aspect framework in producing more coherent and well-separated clusters, offering a more reliable foundation for semantic organization.

\noindent \textbf{Superior structure alignment.}
We achieve the highest CEDS score of $23.8$, indicating strong structural consistency with oracle taxonomies. The \text{HSR} score of $74.5$ confirms that our method possesses the ability to recover coherent hierarchical relations. In addition, the node ratio of $1.2$ suggests a balanced taxonomy size, avoiding the situation of both over-fragmentation and under-segmentation.

\noindent \textbf{Preferred by human evaluators.}
As shown in Table~\ref{tab:main-results}, our method receives the highest human evaluation scores in all five dimensions, with notable improvements in \textbf{Coverage} ($50.6$), \textbf{Structure} ($59.6$), and \textbf{Adequacy} ($54.4$). 
This indicates that our generated taxonomies cover more comprehensive contents and exhibit a more coherent organization of the structure, thereby enhancing the usability.
The agreement between the annotators measured by Fleiss's Kappa on discretized scores (converted from a scale of 1 to 100 to a scale of 5 points) is $0.24$, indicating moderate consistency among the evaluators.

\subsection{Ablation Study on Aspect Generation and Dynamic Search}

\begin{table}[t]
  \centering
\small
    \begin{tabular}{lccccc}
    \toprule
      & \multicolumn{3}{c}{\textbf{Categorization}} & \multicolumn{2}{c}{\textbf{Structure}} \\
      & \textbf{NMI} & \textbf{ARI} & \textbf{Purity} & \textbf{CEDS} & \textbf{HSR} \\
    \midrule
    \multicolumn{4}{l}{\textbf{Dynamic Aspects}} &   &  \\
    \midrule
    Search & 57.8  & 20.1  & 66.4  & 23.7  & 69.9  \\
    \multicolumn{1}{l}{Prune} & 58.6  & 20.4  & 66.0  & 23.9  & 69.4  \\
    \midrule
    \multicolumn{4}{l}{\textbf{Fixed Aspects}}  &   &  \\
    \midrule
    Search & 55.2  & 19.5  & 62.4  & 25.8  & 68.6  \\
    \multicolumn{1}{l}{Prune} & 55.0  & 19.7  & 60.7  & 25.4  & 66.5  \\
    \midrule
    Abstract & 57.1  & 22.3  & 64.3  & 24.2  & 66.3  \\
    \bottomrule
    \end{tabular}%
  \caption{Ablation results on aspect generation and dynamic search. ``Dynamic Aspects'' means our dynamic aspect generation process, while ``Fixed Aspects'' is using fixed manual aspects. ``Search'' denotes dynamic clusters search and ``Prune'' is the pruning strategy in the search process. ``Abstract'' means only using the paper abstracts without aspect guidance.}
   \vspace{-0.2cm}
    \label{tab:ablation}%
\end{table}%

We conduct an ablation study to examine the impact of aspect generation methods and clustering strategies on taxonomy quality in Table~\ref{tab:ablation}.

\noindent \textbf{Dynamic \textit{v.s.} Fixed Aspects.}
We first compare our proposed dynamic aspect generation (\textit{Dynamic Aspects}) with a manually defined aspect template shared across all paper sets (\textit{Fixed Aspects}). The results show that the dynamic aspects achieve consistently better performance in both categorization (\textit{e.g.}, NMI $57.8$ \textit{v.s.} $55.2$) and structural alignment (\textit{e.g.}, HSR $69.9$ \textit{v.s.} $68.6$). This highlights the benefit of tailoring semantic dimensions to each paper set, which better captures latent topical variations and improves clustering quality.

\noindent \textbf{Full \textit{v.s.} Pruning Search.}
Within each setting, we compare two clustering strategies: Full Search and Pruning Search. For the fixed-aspect setting, pruning significantly reduces categorization and structure performance, indicating that simple greedy filtering may break high-quality groupings formed under strong human priors. In contrast, under the dynamic aspect setting, pruning yields comparable performance to full dynamic search. This suggests that while LLM-generated aspects offer higher representational flexibility, they also introduce variability and redundancy, where pruning can help remove outliers with little degradation.

\noindent \textbf{Effect of Using Abstracts Only.}
Finally, we include a baseline that uses only abstracts of papers without aspects. Although it performs reasonably well in ARI (22.3), its overall categorization and structure scores remain lower than our full model. This underscores the importance of aspect-guided representation beyond manual summarization.



\subsection{Effect of Hyperparameter \texorpdfstring{$k_v$}{K}}





\begin{table}[t]
  \centering
    \small
    \begin{tabular}{ccccccc}
    \toprule
    \multirow{2}[2]{*}{$k_v$} & \multicolumn{3}{c}{\textbf{Categorization}} & \multicolumn{2}{c}{\textbf{Structure}} & \multicolumn{1}{c}{\multirow{2}{*}{\textbf{Nodes}}} \\
      & \textbf{NMI} & \textbf{ARI} & \textbf{Purity} & \textbf{CEDS} & \textbf{HSR} &\\
    \midrule
    3 & 55.1  & 21.7  & 60.2  & 24.6  & 63.7  & 1.1  \\
    4 & 57.6  & 19.5  & 65.2  & 24.3  & 69.3  & 1.4  \\
    5 & 59.0  & 18.9  & 69.5  & 20.4  & 68.9  & 1.6  \\
    6 & 61.2  & 18.2  & 73.6  & 19.9  & 69.9  & 1.9  \\
    \midrule
    S & 56.2  & 21.5  & 62.2  & 23.9  & 66.0  & 1.1 \\
    \bottomrule
    \end{tabular}%
    \caption{Performance under different values of hyperparameter $k_v$, which controls the number of clusters per node. ``S'' denotes an adaptive selection strategy from our baseline. Fixed larger $k_v$ improves purity but harms structural consistency (CEDS and Nodes), while adaptive $k_v$ achieves a balanced yet unremarkable performance across all metrics.}
     \vspace{-0.2cm}
    \label{tab:k-selection}%
\end{table}%

We analyze the influence of the hyperparameter $k_v$, which controls the number of clusters generated at each node during hierarchical taxonomy construction. Table~\ref{tab:k-selection} reports the results under fixed values of $k_v \in \{3, 4, 5, 6\}$, as well as an adaptive strategy (``S'') \cite{katz-etal-2024-knowledge} where the model dynamically selects from ${3, 4, 5, 6}$ based on the clustering result with the highest silhouette score.

\noindent \textbf{Fixed \textit{v.s.} Adaptive $k_v$.}
As $k_v$ increases, we observe a steady improvement in categorization performance, with NMI rising from 55.1 (at $k_v=3$) to 61.2 (at $k_v=6$). Purity also increases substantially, reflecting finer-grained clustering. However, this comes at the cost of structural quality: CEDS decline and the normalized node count (Nodes) increase, indicating over-fragmented taxonomies with reduced alignment to the gold standard.


The adaptive strategy achieves relatively balanced performance across all metrics rather than a significant improvement in any individual metric (NMI $56.2$, ARI $21.5$, CEDS $23.9$).
Moreover, the adaptive strategy requires repeated clustering operations for all $k_v$, resulting in substantially higher computational overhead.  
Coupled with only marginal improvements, the high cost suggests that silhouette-based selection may offer limited practical benefit in taxonomy generation.


\tikzset{
    basic/.style  = {draw, text width=2cm, drop shadow, font=\sffamily, rectangle},
    root/.style   = {basic, rounded corners=2pt, thin, align=center,
                     fill=green!20},
    onode/.style = {basic, thin, align=left, fill=pink!10!blue!80!red!10, text width=6.5em},
    onode2/.style = {basic, thin, align=left, fill=pink!10!blue!80!red!10, text width=4.3cm},
    xnode/.style = {basic, thin, align=left, fill=pink!60, text width=4.5cm},
    xnode2/.style = {basic, thin, align=left, fill=pink!60, text width=6cm},
    tnode/.style = {basic, thin, rounded corners=2pt, align=center, fill=blue!20,text width=5cm,},
    tnode2/.style = {basic, thin, rounded corners=2pt, align=center, fill=blue!20,text width=4cm,},
    wnode/.style = {basic, thin, rounded corners=2pt, align=center, fill=black!5,text width=3cm}, 
    wnode2/.style = {basic, thin, rounded corners=2pt, align=center, fill=black!5,text width=5cm}, 
    wnode3/.style = {basic, thin, rounded corners=2pt, align=center, fill=black!5,text width=4cm}, 
    edge from parent/.style={draw=black, edge from parent fork right}
}

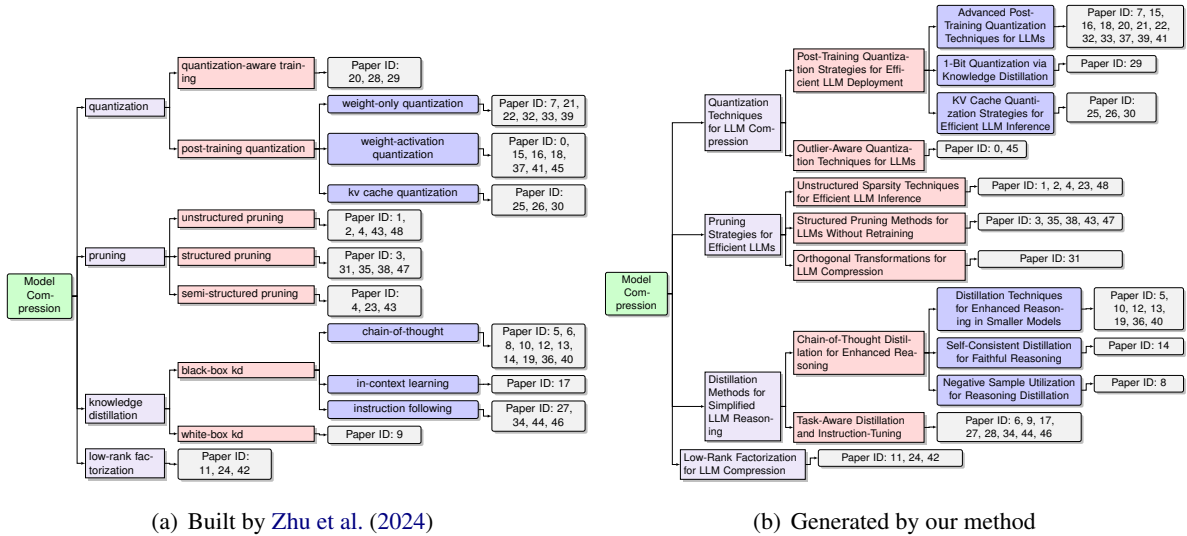
\begin{figure*}[ht]
    \vspace{-0.3cm}
    \centering
\subfigure[Built by \citet{zhu-etal-2024-survey-model}]{
    \begin{minipage}[b]{0.48\textwidth}
          \resizebox{\textwidth}{!}{
            \begin{forest} for tree={
                grow=east,
                growth parent anchor=east,
                parent anchor=east,
                child anchor=west,
                edge path={\noexpand\path[\forestoption{edge},->, >={latex}] 
                     (!u.parent anchor) -- +(5pt,0pt) |- (.child anchor)
                     \forestoption{edge label};}
            }
[Model Compression, root
  [low-rank factorization, onode
    [Paper ID: 11{,} 24{,} 42, wnode]]
  [knowledge distillation, onode
    [white-box kd, xnode
      [Paper ID: 9, wnode]]
    [black-box kd, xnode
      [instruction following, tnode
        [Paper ID: 27{,} 34{,} 44{,} 46, wnode]]
      [in-context learning, tnode
        [Paper ID: 17, wnode]]
      [chain-of-thought, tnode
        [Paper ID: 5{,} 6{,} 8{,} 10{,} 12{,} 13{,} 14{,} 19{,} 36{,} 40, wnode]]]]
  [pruning, onode
    [semi-structured pruning, xnode
      [Paper ID: 4{,} 23{,} 43, wnode]]
    [structured pruning, xnode
      [Paper ID: 3{,} 31{,} 35{,} 38{,} 47, wnode]]
    [unstructured pruning, xnode
      [Paper ID: 1{,} 2{,} 4{,} 43{,} 48, wnode]]]
  [quantization, onode
    [post-training quantization, xnode
      [kv cache quantization, tnode
        [Paper ID: 25{,} 26{,} 30, wnode]]
      [weight-activation quantization, tnode
        [Paper ID: 0{,} 15{,} 16{,} 18{,} 37{,} 41{,} 45, wnode]]
      [weight-only quantization, tnode
        [Paper ID: 7{,} 21{,} 22{,} 32{,} 33{,} 39, wnode]]]
    [quantization-aware training, xnode
      [Paper ID: 20{,} 28{,} 29, wnode]]]]
            \end{forest}
        }
    \vspace{-0.3cm}
    \label{categorization_of_LLMs}
    \end{minipage}\hfill
}
\subfigure[Generated by our method]{
    \begin{minipage}[b]{0.48\textwidth}
        \centering
        \resizebox{\textwidth}{!}{
            \begin{forest} for tree={
                grow=east,
                growth parent anchor=east,
                parent anchor=east,
                child anchor=west,
                edge path={\noexpand\path[\forestoption{edge},->, >={latex}] 
                     (!u.parent anchor) -- +(5pt,0pt) |- (.child anchor)
                     \forestoption{edge label};}
            }
[Model Compression, root
  [Low-Rank Factorization for LLM Compression, onode2
    [Paper ID: 11{,} 24{,} 42, wnode2]]
  [Distillation Methods for Simplified LLM Reasoning, onode
    [Task-Aware Distillation and Instruction-Tuning, xnode
      [Paper ID: 6{,} 9{,} 17{,} 27{,} 28{,} 34{,} 44{,} 46, wnode2]]
    [Chain-of-Thought Distillation for Enhanced Reasoning, xnode      
      [Negative Sample Utilization for Reasoning Distillation, tnode
        [Paper ID: 8, wnode]]
      [Self-Consistent Distillation for Faithful Reasoning, tnode
        [Paper ID: 14, wnode]]
      [Distillation Techniques for Enhanced Reasoning in Smaller Models, tnode
        [Paper ID: 5{,} 10{,} 12{,} 13{,} 19{,} 36{,} 40, wnode]]]]
  [Pruning Strategies for Efficient LLMs, onode    
    [Orthogonal Transformations for LLM Compression, xnode2
      [Paper ID: 31, wnode2]]
    [Structured Pruning Methods for LLMs Without Retraining, xnode2
      [Paper ID: 3{,} 35{,} 38{,} 43{,} 47, wnode2]]
    [Unstructured Sparsity Techniques for Efficient LLM Inference, xnode2
      [Paper ID: 1{,} 2{,} 4{,} 23{,} 48, wnode2]]]
      [Quantization Techniques for LLM Compression, onode
    [Outlier-Aware Quantization Techniques for LLMs, xnode
      [Paper ID: 0{,} 45, wnode]]
    [Post-Training Quantization Strategies for Efficient LLM Deployment, xnode
      [KV Cache Quantization Strategies for Efficient LLM Inference, tnode2
        [Paper ID: 25{,} 26{,} 30, wnode]]
      [1-Bit Quantization via Knowledge Distillation, tnode2
        [Paper ID: 29, wnode]]
    [Advanced Post-Training Quantization Techniques for LLMs, tnode2
        [Paper ID: 7{,} 15{,} 16{,} 18{,} 20{,} 21{,} 22{,} 32{,} 33{,} 37{,} 39{,} 41, wnode3]]]]
    ]
            \end{forest}
        }
    \vspace{-0.3cm}
    \label{categorization_of_LLMs_pred}
    \end{minipage}
    }
    \vspace{-0.2cm}
    \caption{Taxonomy of "Model Compression methods for Large Language Models".}
     \vspace{-0.3cm}
    \label{fig:case-study}
\end{figure*}

\subsection{Robustness to Noisy Inputs}

To evaluate the robustness of our method under more realistic settings, where the initial set of relevant papers is not perfectly curated, we conducted additional experiments simulating noisy input conditions, as suggested by Reviewer RHVv. Specifically, we injected 5\%–30\% unrelated papers into the curated dataset to mimic potential noise introduced by retrieval-based pipelines.

\begin{table}[htbp]
  \centering
 \small
 \setlength{\tabcolsep}{2.5pt}
    \begin{tabular}{ccrrrrrc}
    \toprule
      & \multicolumn{1}{c}{\multirow{2}[1]{*}{\shortstack{\textbf{Noise}\\\textbf{Ratio}}}} & \multicolumn{3}{c}{\textbf{Categorization}} & \multicolumn{2}{c}{\textbf{Structure}} & \multicolumn{1}{c}{\multirow{2}[2]{*}{\textbf{Nodes}}} \\
      &   & \multicolumn{1}{c}{\textbf{NMI}} & \multicolumn{1}{c}{\textbf{ARI}} & \multicolumn{1}{c}{\textbf{Purity}} & \multicolumn{1}{c}{\textbf{CEDS}} & \multicolumn{1}{c}{\textbf{HSR}} &  \\
    \midrule
    \multirow{5}[2]{*}{\shortstack{\textbf{TnT-} \\ \textbf{LLM}}} & 0\% & 43.76  & 3.61  & 56.16  & 20.06  & 72.15  & 1.66\\
      & 5\% & 44.68  & 3.91  & 56.26  & 20.87  & 75.48  & 1.92  \\
      & 10\% & 50.73  & 3.88  & 63.45  & 15.88  & 81.57  & 2.36  \\
      & 20\% & 41.25  & 1.90  & 52.42  & 17.71  & 72.58  & 1.79  \\
      & 30\% & 43.55  & 3.43  & 55.12  & 19.10  & 76.54  & 2.20  \\
    \midrule
    \multirow{5}[2]{*}{\shortstack{\textbf{SCYC} \\ \textbf{HIC}}} & 0\% & 39.37  & 6.25  & 45.29  & 17.36  & 60.52  & 1.73 \\
      & 5\% & 38.33  & 5.89  & 44.95  & 16.92  & 61.47  & 1.73  \\
      & 10\% & 37.22  & 5.18  & 46.20  & 15.42  & 63.10  & 1.73  \\
      & 20\% & 37.62  & 5.59  & 43.70  & 17.78  & 62.26  & 1.73  \\
      & 30\% & 36.47  & 6.49  & 46.89  & 17.19  & 63.58  & 1.78 \\
    \midrule
    \multirow{5}[2]{*}{\textbf{Ours}} & 0\% & 53.80  & 13.94  & 61.93  & 23.99  & 68.28  & 1.48 \\
      & 5\% & 54.46  & 15.74  & 62.93  & 23.15  & 69.30  & 1.49  \\
      & 10\% & 52.93  & 14.02  & 61.05  & 23.22  & 70.22  & 1.65  \\
      & 20\% & 54.17  & 16.54  & 61.33  & 23.72  & 70.69  & 1.73  \\
      & 30\% & 54.42  & 17.53  & 61.64  & 21.66  & 72.14  & 1.80  \\
    \bottomrule
    \end{tabular}%
 \caption{Performance comparison under different noise levels.}

  \label{tab:noise}%
\end{table}%

Experimental results (see Table \ref{tab:noise}) show that our method consistently outperforms baseline approaches and maintains superior performance and structural stability across all noise levels. 
In contrast, TnT-LLM suffers from significant performance fluctuations, and SCYCHIC experiences moderate degradation. 

We attribute this robustness to two key design choices in our framework:
Aspect-aware clustering with dynamic search, which selectively identifies the most relevant combination of aspect dimensions for each paper, effectively filtering out noise;
Expanded representation space of aspect-cluster combinations, which allows noisy or outlier papers to be isolated into peripheral nodes without disrupting the core taxonomy structure.

These findings highlight the error-tolerant nature of our approach and demonstrate its effectiveness even when applied to noisy, less curated document sets. We believe this provides strong evidence of the method’s practical applicability beyond oracle-like experimental conditions.

\subsection{Case-Study}

\noindent \textbf{Comparison with Human-Annotated Taxonomy.}
Figure~\ref{fig:case-study}(a) shows the human-annotated taxonomy from \citet{zhu-etal-2024-survey-model} on “Model Compression Methods for Large Language Models,” and Figure~\ref{fig:case-study}(b) presents our generated result.
For comparison, additional case studies produced by baseline methods are included in the Appendix \ref{sec:case-taxo}
At the top level, both taxonomies adopt a method-based categorization (\textit{e.g.}, quantization, pruning, distillation), which is largely consistent. Only one paper (28) is misclassified.
In deeper layers, our taxonomy introduces more fine-grained and diverse subtopics. While these differ from the human taxonomy, they reflect alternative yet valid grouping strategies based on implementation details or use cases. This highlights the subjectivity of deeper-level structuring and the model's ability to surface meaningful semantic distinctions.

%% file: related.tex
\section{Related Work}

Organizing the ever-growing scientific literature into coherent, hierarchical categories remains a core challenge in scholar knowledge management.
Traditional approaches typically rely on manually curated taxonomies, where each paper is mapped to one or more predefined categories within a multi-level hierarchy \cite{zhang2021hierarchical, sadat-caragea-2022-hierarchical, rao2023hierarchical}. 

Recent advances in LLMs have significantly reshaped the landscape of topic modeling and document clustering by semantically rich and context-aware representations, allowing for more interpretable and scalable taxonomy construction \cite{zhang2023clusterllm,pham2024topicgpt,wang2023prompting,qiu2024topic,viswanathan_LargeLanguageModels_2024}.
In general, there are two technical paradigms for taxonomy construction: \textbf{classification-based} and \textbf{clustering-based}, where each of them offers distinct advantages and trade-offs.

In the classification paradigm, an LLM first induces a taxonomy, and papers are subsequently assigned \cite{pham2024topicgpt}. 
CHIME \cite{hsu2024chime} produces the taxonomy and assigns papers in one pass. 
GoalEx \cite{wang2023goal} aligns LLM-generated explanations with papers and applies integer linear programming to finalize a non-overlapping set of assignments.
To better handle long-document settings, TnT-LLM \cite{wan_TnTLLMTextMining_2024} iteratively generates and updates the label taxonomy. 
More recently, TaxoAdapt \cite{kargupta-etal-2025-taxoadapt} incrementally expands the taxonomy by analyzing papers one by one, formulating multi-dimensional taxonomy construction as iterative multi-label classification.
Automatic literature review generation pipelines such as AutoSurvey \cite{2024autosurvey}, Storm \cite{shao-etal-2024-assisting}, and SurveyForge \cite{yan-etal-2025-surveyforge} replace taxonomies with hierarchically structured outlines, i.e., they first draft an outline of target topics and then retrieve and attach relevant papers to each entry.

Despite their flexibility, label-first pipelines often produce redundant labels, hallucinated or missing categories, and imbalanced hierarchies.
\textbf{Clustering-based methods} organize papers in the representation space, then leverage inter-paper relations to enforce global coherence and balance, and finally add labels respectively.
A related line integrates clustering with LLM generation, where papers are first grouped by unsupervised methods and then semantic labels are produced for each cluster \cite{diaz2025k,hu2024taxonomy}. 
Knowledge Navigator \cite{katz-etal-2024-knowledge} performs single-stage flat clustering, while \citet{gao_ScienceHierarchographyHierarchical_2025} explore hierarchical strategies (bottom-up, top-down, and bi-direction).   
However, these approaches often rely on local, per-cluster descriptions in isolation, yielding redundant or inconsistent labels due to missing global context and weak structural constraints.
In contrast, our method proposes dynamic and structure-aware hierarchical clustering with global aspects, maintaining the semantic distinctiveness and structural fidelity of the taxonomies.

%% file: limitation.tex
\section*{Limitations}

Although our method demonstrates strong performance, several limitations remain:
\begin{enumerate}
    \item In practical applications, the system must first retrieve candidate papers from a broad and potentially noisy corpus, which introduces additional challenges such as incomplete coverage, irrelevant documents, and retrieval errors. Our framework focuses on a controllable experimental environment with oracle papers. Developing retrieval-integrated taxonomy construction methods that are robust to these issues constitutes an important direction for future work.
    \item The quality of aspect extraction and summarization depends on the capabilities of the underlying LLM, which affects the generalization of our framework.
    \item The combination of multi-aspect encoding and iterative clustering introduces computational overhead, which may limit scalability to very large corpora. We plan to explore more efficient clustering strategies and scalable approximations to support deployment on a greater scale.
    \item Our evaluation benchmark focuses on survey papers in computer science, where its applicability to other domains or less-structured corpora remains to be explored. In future work, we will extend our framework to cross-domain settings.
    \item We find that silhouette-based $k$-selection is not well suited for clustering in complex and semantic-driven tasks such as taxonomy generation, which leaves the development of more effective task-specific clustering selection strategies for future work.
    \item Our current framework employs hierarchical clustering, which enforces a strict, non-overlapping partitioning of papers at each level. In contrast, expert-authored taxonomies (e.g., the oracle trees in our benchmark) sometimes allow a paper to be assigned to multiple branches. Enabling multi-label taxonomy construction is thus an important and challenging extension that we leave for future research.
\end{enumerate}

\section*{Acknowledgments}
This work was supported by the National Natural Science Foundation of China (NSFC) (grant 62276078, U22B2059), the Key R\&D Program of Heilongjiang via grant 2022ZX01A32, and the Fundamental Research Funds for the Central Universities (XNJKKGYDJ2024013).  It was also supported by the Ministry of Education, Singapore, under its AcRF Tier 2 Funding (Proposal ID: T2EP20123-0052). Any opinions, findings and conclusions or recommendations expressed in this material are those of the author(s) and do not reflect the views of the Ministry of Education, Singapore. We thank the iFLYTEK Spark AI Assistant Team for providing application requirements and high-value feedback.

\section*{Ethics Statement}

This work focuses on constructing paper taxonomies using large language models (LLMs), with the goal of assisting researchers and beginners in understanding domain knowledge, tracking research trends, and improving reading efficiency. While this technology has the potential to support scientific discovery and education, it also carries risks that warrant ethical consideration.

\noindent \textbf{Use of LLMs and Potential Risks}
Our framework relies on LLMs to generate semantic aspects and organize papers into a hierarchical taxonomy. We acknowledge that LLMs are susceptible to hallucinations, which may lead to factually incorrect or misleading taxonomy structures. Nevertheless, any downstream use of the generated taxonomy for scientific analysis or educational purposes should be critically verified, especially in high-stakes or sensitive applications.

\noindent \textbf{Dataset Collection and Licensing}
We construct our dataset using publicly available metadata and content from \href{https://arxiv.org/}{arXiv} and \href{https://www.semanticscholar.org/}{Semantic Scholar}, both of which provide research access under open licenses. The dataset used in this study includes paper titles, metadata (e.g., authors, publication years), and taxonomy structures extracted from the \LaTeX\ source files of review papers collected from arXiv. Specifically, we target survey papers that explicitly include taxonomy structures in their source files. From these files, we extract the taxonomy tree as well as the titles of cited papers mentioned within the taxonomy.

For each cited paper in the taxonomy, we obtain its metadata using the Semantic Scholar API. In cases where the cited papers are also publicly available on arXiv, we further retrieve their \LaTeX\ source files and extract their \texttt{Introduction} sections. This allows us to enrich the representation of each paper beyond the abstract and metadata, enabling more informed and semantically grounded taxonomy construction.

All data were obtained through open APIs and publicly accessible sources, and their use is restricted to academic research. We confirm that our use of these artifacts complies with their intended use and access conditions. No redistribution of full-text content outside permitted use cases has been conducted. The resulting dataset, including derived taxonomy annotations, is shared under a research-only license and should not be repurposed for commercial or non-academic use.

\noindent \textbf{Privacy and Anonymization}
We conducted a manual check to ensure that the dataset does not contain personally identifiable information (PII) beyond standard academic author metadata, which are already publicly accessible through the original platforms. No sensitive personal content, user-generated data, or non-consensual information is included. Our system does not process or generate user data, and all derived outputs (e.g., cluster labels, taxonomy facets) are generated from published research papers.

\noindent \textbf{Human Annotation and Consent}
We recruited voluntary annotators to evaluate the quality of the generated taxonomies. All annotators were fully informed about the purpose of the study, the nature of the data, and how their assessments would be used. No personal information was collected from annotators, and consent was obtained prior to their participation.

%% file: appendix.tex
\section{Evaluation Metrics}

\label{sec:appendix-metrics}









We evaluate taxonomy generation from two complementary perspectives: \textit{clustering structure} and \textit{heading quality}. In addition to automatic evaluation, we also conduct human evaluation to assess the practical quality of the generated taxonomies.

\subsection{Clustering Evaluation}

\textbf{Hierarchical Mutual Information (HMI)} extends mutual information to hierarchical structures by evaluating consistency across multiple levels of the taxonomy. It provides a structure-aware measure that rewards alignment not only at the leaf level but also across internal nodes.

\noindent \textbf{Adjusted Rand Index (ARI)} measures the agreement between the predicted and gold cluster assignments, correcting for random chance. It is widely used in clustering evaluation and is robust to varying cluster sizes.

\noindent \textbf{Purity} quantifies the extent to which each predicted cluster contains documents from a single ground-truth category. While intuitive, this metric may favor solutions with a large number of small clusters.

\subsection{Heading Evaluation}

\textbf{Heading Soft Recall}. We follow the calculation of \citet{shao-etal-2024-assisting}.
This metric measures the proportion of ground-truth headings that are approximately matched by generated node names using soft string similarity. It allows for minor lexical variations and captures semantic overlap.
It is worth noting that, in theory, longer generated outputs tend to achieve higher scores under soft matching metrics such as Soft Heading Recall. This is because longer outputs are more likely to semantically overlap with the reference headings, thereby increasing the chance of a successful match under relaxed similarity thresholds. However, this improvement may not necessarily reflect better quality, as it can be attributed to over-generation rather than more accurate content selection.

\noindent \textbf{Catalogue Edit Distance Similarity (CEDS)} \cite{zhu-etal-2023-hierarchical} evaluates the overall similarity between the generated taxonomy and the gold taxonomy by computing a \textbf{normalized tree edit distance}. It accounts for both structural alignment (e.g., insertion, deletion, reordering of nodes) and heading-level similarity, offering a holistic assessment of taxonomy quality.

\subsection{Human Evaluation}

To complement automatic metrics, we conduct a human evaluation based on five criteria followed \citet{hu_TaxonomyTreeGeneration_2025}:

\begin{itemize}
\item \textbf{Coverage}: Does the taxonomy comprehensively cover the major themes and subtopics within the document collection?
\item \textbf{Relevance}: Are the identified categories appropriate and meaningful for the given set of documents?
\item \textbf{Structure}: Is the overall organization coherent and logically structured as a hierarchy?
\item \textbf{Usefulness}: How helpful is the taxonomy for readers trying to understand or navigate the domain?
\item \textbf{Validity}: Does the taxonomy align with expert expectations or established domain knowledge?
\end{itemize}


Each aspect is rated on a scale of $1$ to $100$ by multiple annotators with relevant domain expertise, and the final scores are averaged among the raters.  
To link the evaluation protocol with concrete outcomes, we further analyze inter-rater reliability by discretizing the scores into five bins of equal width and computing consistency both within and across rater groups.  
Inter-annotator agreement, measured by Fleiss’ $\kappa$ on the discretized ratings, shows the following: Human--Human $=0.31$, LLM--LLM $=0.38$, and Human--LLM $=0.24$.  
Taken together, these results indicate that both human annotators and LLMs exhibit comparable levels of consistency within the group, while their agreement between groups remains relatively low, suggesting systematic differences in rating behavior between the two.

\begin{figure*}[ht]
    \centering
    \includegraphics[width=1\linewidth]{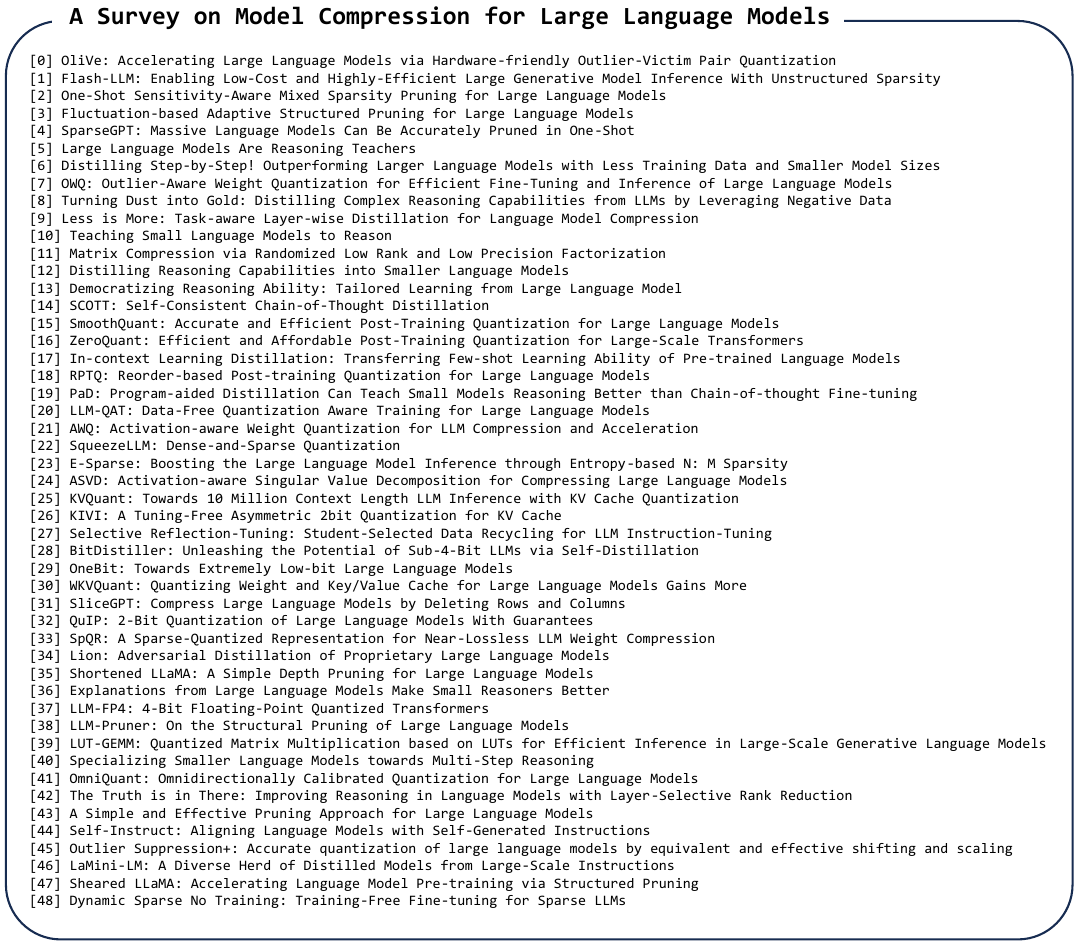}
    \caption{Papers in the taxonomy built by \citet{zhu-etal-2024-survey-model}}
    \label{fig:survey_papers}
\end{figure*}

\section{Case Study}

To qualitatively evaluate the effectiveness of our method, we conduct a case study on the topic of ”Model Compression". 

\subsection{Taxonomy Trees}
\label{sec:case-taxo}

\begin{figure}[!ht]
    \centering
    \includegraphics[width=1\linewidth]{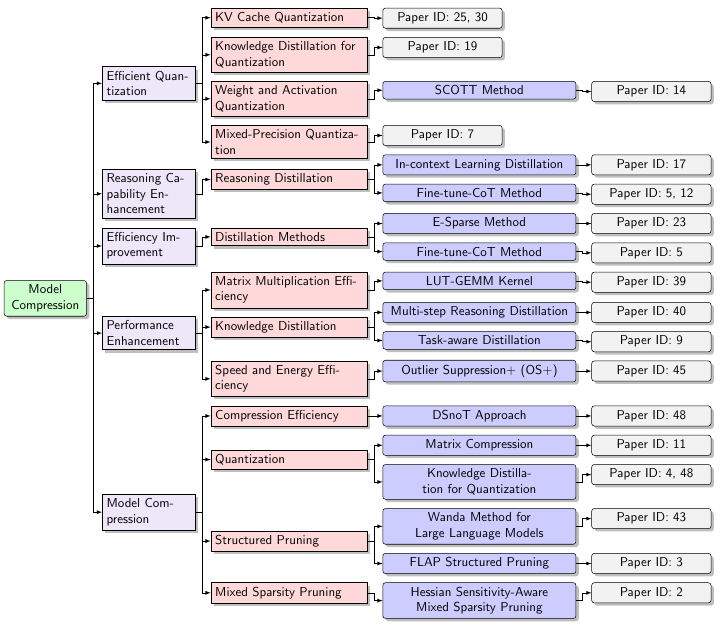}
    \caption{Taxonomy of "Model Compression methods for Large Language Models" generated by Chime \cite{hsu2024chime}.}
    \label{categorization_of_LLMs-chime}
\end{figure}

\begin{figure}[!ht]
    \centering
    \includegraphics[width=1\linewidth]{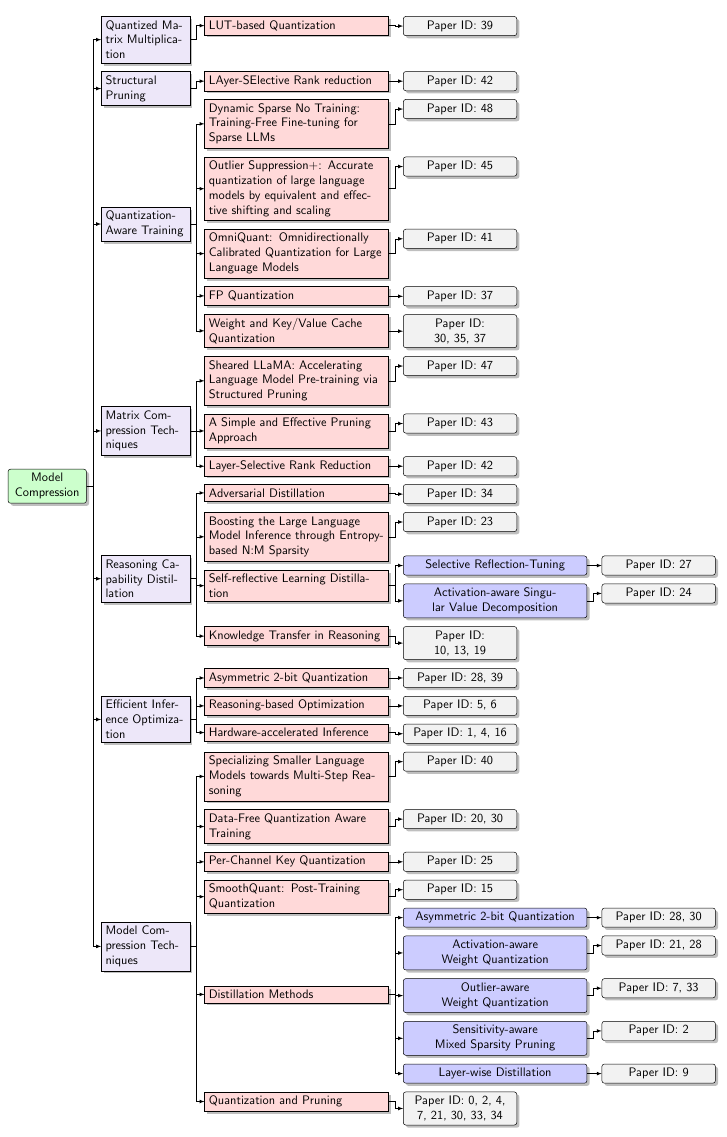}
    \caption{Taxonomy of "Model Compression methods for Large Language Models" generated by TnTLLM \cite{wan_TnTLLMTextMining_2024}.}
    \label{categorization_of_LLMs_tntllm}
\end{figure}

\begin{figure}[!ht]
     \centering
    \includegraphics[width=1\linewidth]{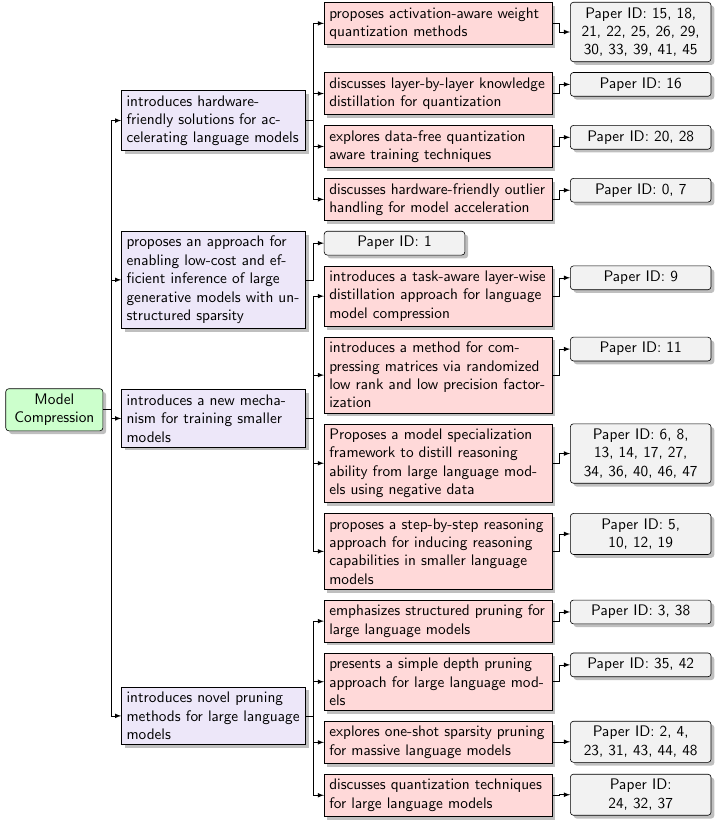}
    \caption{Taxonomy of "Model Compression methods for Large Language Models" generated by GoalEx \cite{wang2023goal}.}
    \label{categorization_of_LLMs_goalex}
\end{figure}

\begin{figure}[!ht]
    \centering
    \includegraphics[width=1\linewidth]{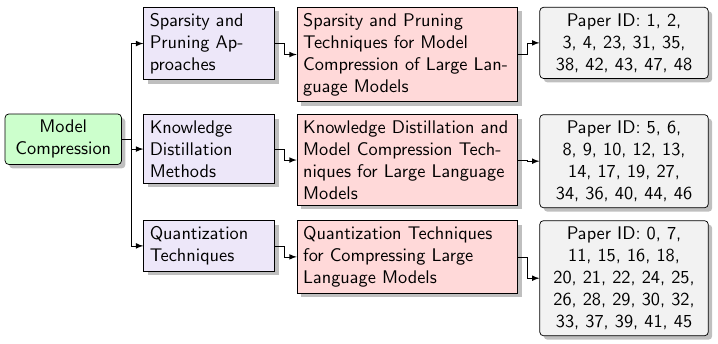}
    \caption{Taxonomy of "Model Compression methods for Large Language Models" generated by KnowledgeNavigator \cite{katz-etal-2024-knowledge}.}
    \label{categorization_of_LLMs_kn}
\end{figure}

\begin{figure}[!ht]
     \centering
    \includegraphics[width=1\linewidth]{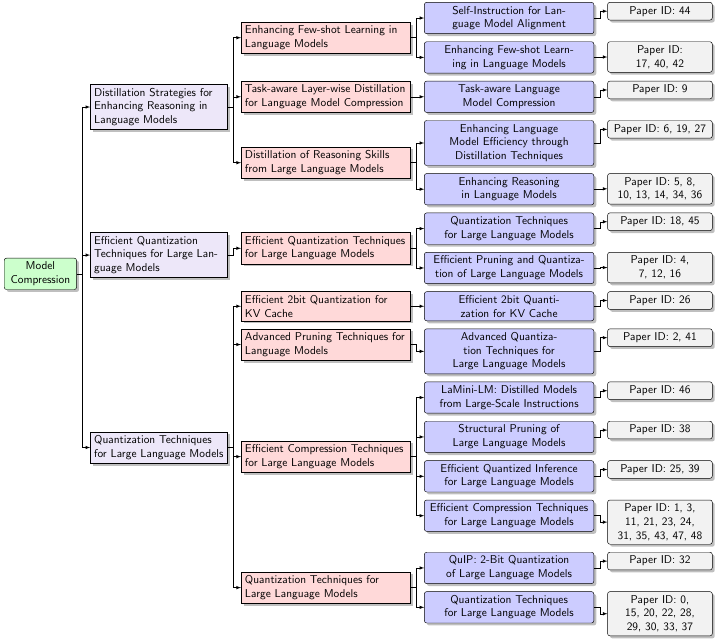}
    \caption{Taxonomy of "Model Compression methods for Large Language Models" generated by SCYCHIC \cite{gao_ScienceHierarchographyHierarchical_2025}.}
    \label{categorization_of_LLMs_scy}
\end{figure}

Figures~\ref{subfig:categorization_of_LLMs_ground} and~\ref{fig:survey_papers} show the human-authored taxonomy tree and the corresponding set of papers from the survey paper "A Survey on Model Compression for Large Language Models"~\cite{zhu-etal-2024-survey-model}. Our generated taxonomy is presented in Figure~\ref{subfig:categorization_of_LLMs_pred}, while the taxonomies produced by other baseline methods are shown in Figures~\ref{categorization_of_LLMs-chime}–\ref{categorization_of_LLMs_scy}.
As illustrated, our method produces a more coherent and semantically meaningful taxonomy structure, with clearer topic hierarchies and better alignment to the source papers, compared to other approaches. 


\tikzset{
    basic/.style  = {draw, text width=2cm, drop shadow, font=\sffamily, rectangle},
    root/.style   = {basic, rounded corners=2pt, thin, align=center,
                     fill=green!20},
    onode/.style = {basic, thin, align=left, fill=pink!10!blue!80!red!10, text width=6.5em},
    onode2/.style = {basic, thin, align=left, fill=pink!10!blue!80!red!10, text width=4.3cm},
    xnode/.style = {basic, thin, align=left, fill=pink!60, text width=4.5cm},
    xnode2/.style = {basic, thin, align=left, fill=pink!60, text width=6cm},
    tnode/.style = {basic, thin, rounded corners=2pt, align=center, fill=blue!20,text width=5cm,},
    tnode2/.style = {basic, thin, rounded corners=2pt, align=center, fill=blue!20,text width=4cm,},
    wnode/.style = {basic, thin, rounded corners=2pt, align=center, fill=black!5,text width=3cm}, 
    wnode2/.style = {basic, thin, rounded corners=2pt, align=center, fill=black!5,text width=5cm}, 
    wnode3/.style = {basic, thin, rounded corners=2pt, align=center, fill=black!5,text width=4cm}, 
    edge from parent/.style={draw=black, edge from parent fork right}
}

\begin{figure*}[ht]
    \centering
\subfigure[Built by \citet{zhu-etal-2024-survey-model}]{
    \begin{minipage}[b]{0.48\textwidth}
          \resizebox{\textwidth}{!}{
            \begin{forest} for tree={
                grow=east,
                growth parent anchor=east,
                parent anchor=east,
                child anchor=west,
                edge path={\noexpand\path[\forestoption{edge},->, >={latex}] 
                     (!u.parent anchor) -- +(5pt,0pt) |- (.child anchor)
                     \forestoption{edge label};}
            }
[Model Compression, root
  [low-rank factorization, onode
    [Paper ID: 11{,} 24{,} 42, wnode]]
  [knowledge distillation, onode
    [white-box kd, xnode
      [Paper ID: 9, wnode]]
    [black-box kd, xnode
      [instruction following, tnode
        [Paper ID: 27{,} 34{,} 44{,} 46, wnode]]
      [in-context learning, tnode
        [Paper ID: 17, wnode]]
      [chain-of-thought, tnode
        [Paper ID: 5{,} 6{,} 8{,} 10{,} 12{,} 13{,} 14{,} 19{,} 36{,} 40, wnode]]]]
  [pruning, onode
    [semi-structured pruning, xnode
      [Paper ID: 4{,} 23{,} 43, wnode]]
    [structured pruning, xnode
      [Paper ID: 3{,} 31{,} 35{,} 38{,} 47, wnode]]
    [unstructured pruning, xnode
      [Paper ID: 1{,} 2{,} 4{,} 43{,} 48, wnode]]]
  [quantization, onode
    [post-training quantization, xnode
      [kv cache quantization, tnode
        [Paper ID: 25{,} 26{,} 30, wnode]]
      [weight-activation quantization, tnode
        [Paper ID: 0{,} 15{,} 16{,} 18{,} 37{,} 41{,} 45, wnode]]
      [weight-only quantization, tnode
        [Paper ID: 7{,} 21{,} 22{,} 32{,} 33{,} 39, wnode]]]
    [quantization-aware training, xnode
      [Paper ID: 20{,} 28{,} 29, wnode]]]]
            \end{forest}
        }
    \label{subfig:categorization_of_LLMs_ground}
    \end{minipage}\hfill
    \vspace{-0.3cm}
}
\subfigure[Generated by our method]{
    \begin{minipage}[b]{0.48\textwidth}
        \centering
        \resizebox{\textwidth}{!}{
            \begin{forest} for tree={
                grow=east,
                growth parent anchor=east,
                parent anchor=east,
                child anchor=west,
                edge path={\noexpand\path[\forestoption{edge},->, >={latex}] 
                     (!u.parent anchor) -- +(5pt,0pt) |- (.child anchor)
                     \forestoption{edge label};}
            }
[Model Compression, root
  [Low-Rank Factorization for LLM Compression, onode2
    [Paper ID: 11{,} 24{,} 42, wnode2]]
  [Distillation Methods for Simplified LLM Reasoning, onode
    [Task-Aware Distillation and Instruction-Tuning, xnode
      [Paper ID: 6{,} 9{,} 17{,} 27{,} 28{,} 34{,} 44{,} 46, wnode2]]
    [Chain-of-Thought Distillation for Enhanced Reasoning, xnode      
      [Negative Sample Utilization for Reasoning Distillation, tnode
        [Paper ID: 8, wnode]]
      [Self-Consistent Distillation for Faithful Reasoning, tnode
        [Paper ID: 14, wnode]]
      [Distillation Techniques for Enhanced Reasoning in Smaller Models, tnode
        [Paper ID: 5{,} 10{,} 12{,} 13{,} 19{,} 36{,} 40, wnode]]]]
  [Pruning Strategies for Efficient LLMs, onode    
    [Orthogonal Transformations for LLM Compression, xnode2
      [Paper ID: 31, wnode2]]
    [Structured Pruning Methods for LLMs Without Retraining, xnode2
      [Paper ID: 3{,} 35{,} 38{,} 43{,} 47, wnode2]]
    [Unstructured Sparsity Techniques for Efficient LLM Inference, xnode2
      [Paper ID: 1{,} 2{,} 4{,} 23{,} 48, wnode2]]]
      [Quantization Techniques for LLM Compression, onode
    [Outlier-Aware Quantization Techniques for LLMs, xnode
      [Paper ID: 0{,} 45, wnode]]
    [Post-Training Quantization Strategies for Efficient LLM Deployment, xnode
      [KV Cache Quantization Strategies for Efficient LLM Inference, tnode2
        [Paper ID: 25{,} 26{,} 30, wnode]]
      [1-Bit Quantization via Knowledge Distillation, tnode2
        [Paper ID: 29, wnode]]
    [Advanced Post-Training Quantization Techniques for LLMs, tnode2
        [Paper ID: 7{,} 15{,} 16{,} 18{,} 20{,} 21{,} 22{,} 32{,} 33{,} 37{,} 39{,} 41, wnode3]]]]
    ]
            \end{forest}
        }
    \label{subfig:categorization_of_LLMs_pred}
    \end{minipage}
    \vspace{-0.3cm}
    }
    \vspace{-0.2cm}
    \caption{Taxonomy of "Model Compression methods for Large Language Models".}
     \vspace{-0.2cm}
    \label{appfig:case-study}
\end{figure*}

\subsection{Generation Process}
\label{sec:case-generation-process}


To complement the quantitative results in Table~\ref{tab:ablation}, we provide several representative case studies that qualitatively illustrate the role of aspect generation, aspect-guided summarization, and facet identification. These examples highlight how different components contribute to clustering outcomes and taxonomy construction.

\begin{figure}[!ht]
     \centering
    \includegraphics[width=1\linewidth]{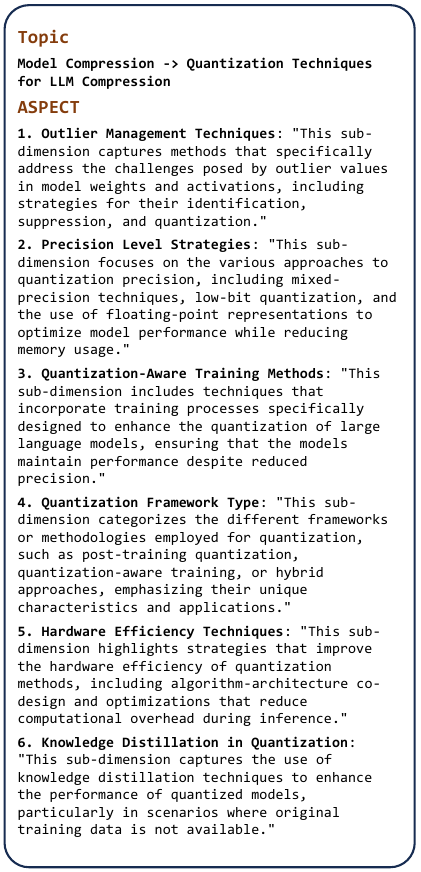}
    \caption{Aspect generation under a specific topic.}
    \label{fig:generation_process_case_aspect}
\end{figure}

Figure~\ref{fig:generation_process_case_aspect} shows the aspects generated under the topic \emph{``Model Compression $\rightarrow$ Quantization Techniques for LLM Compression''}.
The resulting aspects capture salient semantic dimensions that effectively characterize and differentiate relevant papers (e.g., Quantization Framework Type, Hardware Efficiency Techniques).

\begin{figure}[!ht]
     \centering
    \includegraphics[width=1\linewidth]{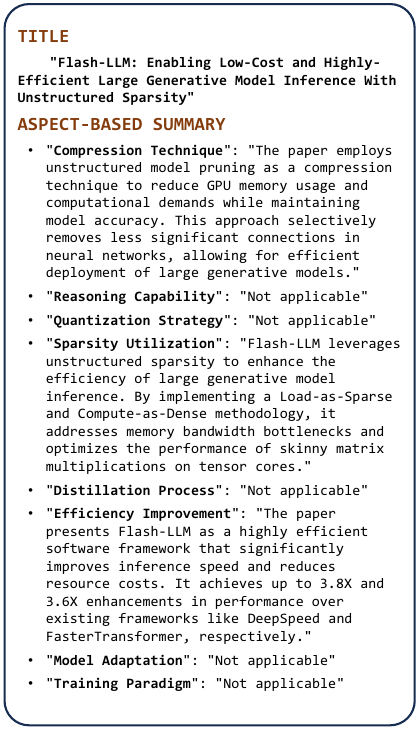}
    \caption{Aspect-based summary generated for the paper \emph{``Flash-LLM: Enabling Low-Cost and Highly-Efficient Large Generative Model Inference With Unstructured Sparsity''}.}
    \label{fig:generation_process_case_summary}
\end{figure}

These aspects are then used in parallel to guide the encoding of individual papers.
Figure~\ref{fig:generation_process_case_summary} presents the aspect-guided summary for the paper \emph{``Flash-LLM: Enabling Low-Cost and Highly-Efficient Large Generative Model Inference With Unstructured Sparsity''}.
Compared with the original abstract, the aspect-based summary selectively foregrounds details aligned with the identified aspects, facilitating clearer alignment with clustering.

\begin{figure}[!ht]
     \centering
    \includegraphics[width=1\linewidth]{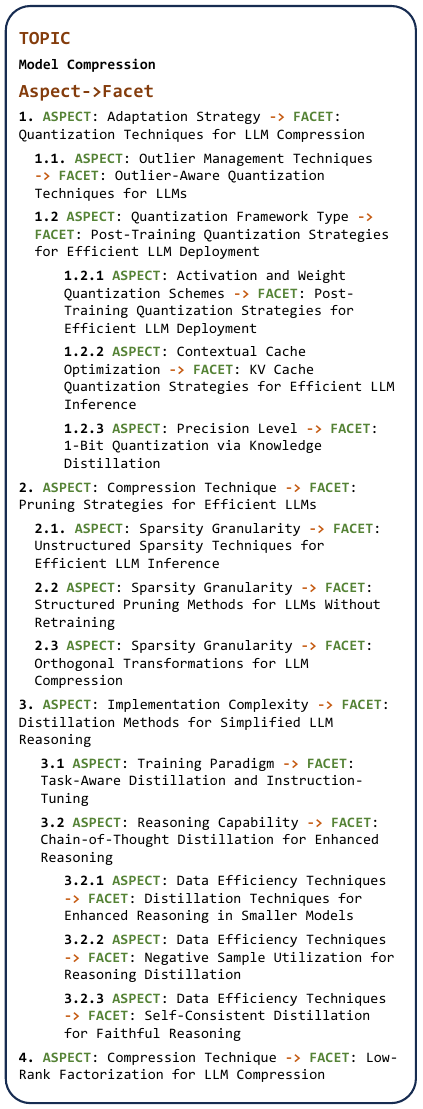}
    \caption{Facet identification within a topic.}
    \label{fig:generation_process_case_facet}
\end{figure}

Figure~\ref{fig:generation_process_case_facet} illustrates how facets are identified within the topic \emph{``Model Compression''} and the selected aspects after clustering with dynamic search.
The system generates corresponding topic facets that summarize the semantic focus of each substructure and render the resulting taxonomy more interpretable and navigable.

\section{Prompts}
 \label{sec:prompts}

\begin{figure}[!ht]
    \centering
    \includegraphics[width=1\linewidth]{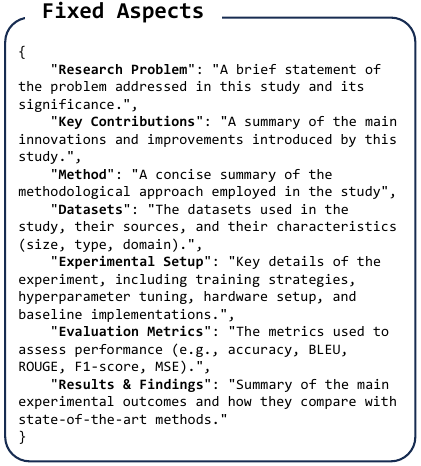}
    \caption{Fixed aspects we used.}
    \label{prompt_fix_aspects}
\end{figure}

\begin{figure}[!ht]
    \centering
    \includegraphics[width=1\linewidth]{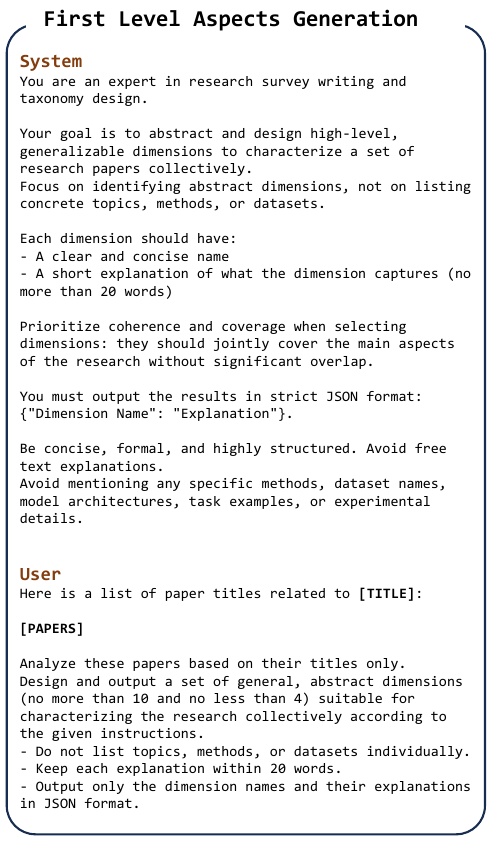}
    \caption{Prompt used for the first-level aspects generation.}
    \label{prompt_high_aspects}
\end{figure}

\begin{figure}[!ht]
    \centering
    \includegraphics[width=1\linewidth]{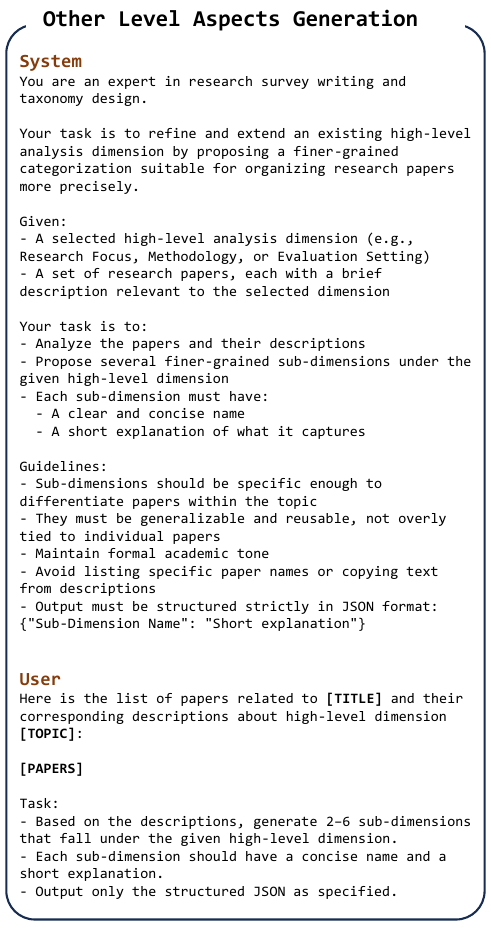}
    \caption{Prompt used for the other-level aspects generation.}
    \label{prompt_specific_aspects}
\end{figure}

\begin{figure}[!ht]
    \centering
    \includegraphics[width=1\linewidth]{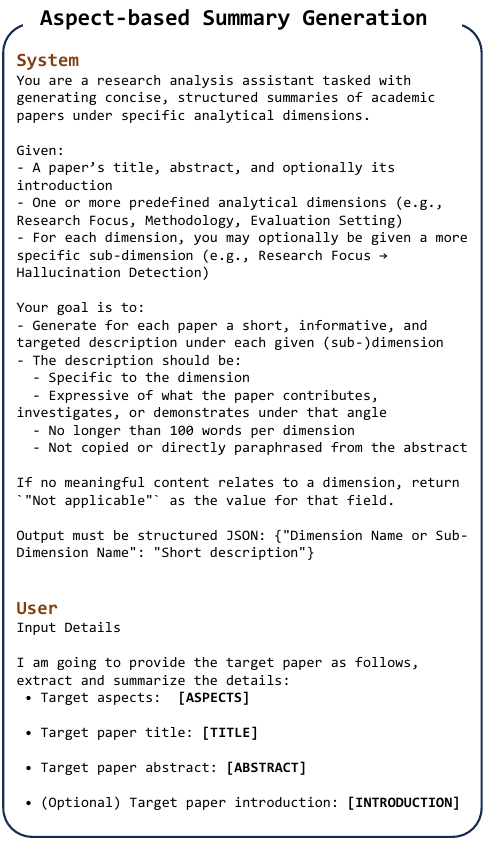}
    \caption{Prompt used for aspect-based summary generation.}
    \label{prompt_aspect_summary}
\end{figure}

\begin{figure}[!ht]
    \centering
    \includegraphics[width=1\linewidth]{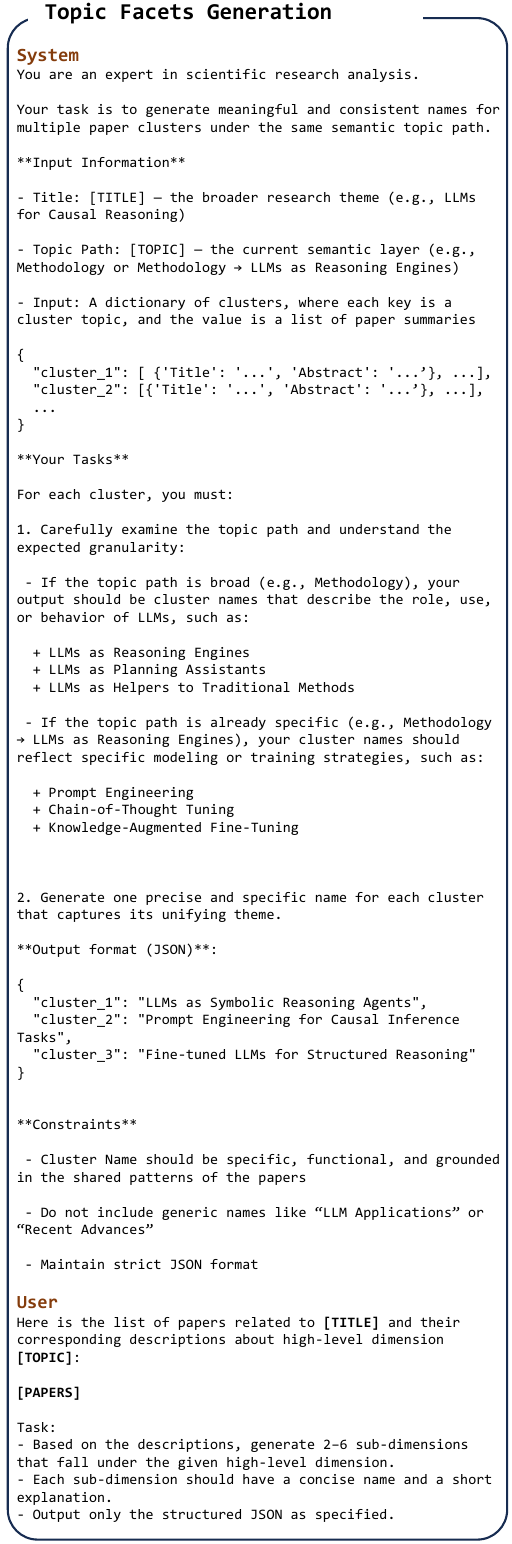}
    \caption{Prompt used for topic facets generation.}
    \label{prompt_heading}
\end{figure}

The prompts we used are shown in Figures~\ref{prompt_fix_aspects}–\ref{prompt_heading}.

